\documentclass{article}

\usepackage{microtype}
\usepackage{graphicx}
\usepackage{wrapfig}
\usepackage{afterpage}
\usepackage{subcaption}
\usepackage{booktabs} 

\usepackage{xcolor} 

\usepackage{forest}
\usepackage{bold-extra}

\usepackage[preprint, nonatbib]{neurips_2026}

\usepackage[
  backend=biber,
  style=authoryear,
  maxcitenames=2,    
  maxbibnames=99,    
  giveninits=true,   
  natbib=true,       
  uniquename=false,
  uniquelist=false,
]{biblatex}
\DeclareCiteCommand{\textcite}
  {\boolfalse{cbx:parens}}
  {\usebibmacro{citeindex}%
   \printtext[bibhyperref]{\usebibmacro{textcite}}}
  {\ifbool{cbx:parens}
     {\bibcloseparen\global\boolfalse{cbx:parens}}
     {}%
   \multicitedelim}
  {\usebibmacro{textcite:postnote}}
\DeclareCiteCommand{\parencite}[\mkbibparens]
  {\usebibmacro{prenote}}
  {\usebibmacro{citeindex}%
   \printtext[bibhyperref]{\usebibmacro{cite}}}
  {\multicitedelim}
  {\usebibmacro{postnote}}
\DeclareCiteCommand{\citealp}
  {\usebibmacro{prenote}}
  {\usebibmacro{citeindex}%
   \printtext[bibhyperref]{\usebibmacro{cite}}}
  {\multicitedelim}
  {\usebibmacro{postnote}}
\renewbibmacro*{cite:labelyear+extrayear}{%
  \iffieldundef{labelyear}
    {}
    {\printfield{labelyear}%
     \printfield{extrayear}}}
\renewbibmacro*{cite}{%
  \iffieldundef{shorthand}
    {\printnames{labelname}%
     \setunit{\printdelim{nameyeardelim}}%
     \iffieldundef{labelyear}
       {}
       {\printfield{labelyear}%
        \printfield{extrayear}}}
    {\usebibmacro{cite:shorthand}}}
\renewbibmacro*{textcite}{%
  \iffieldundef{shorthand}
    {\printnames{labelname}%
     \setunit{\printdelim{nameyeardelim}}%
     \printtext[parens]{%
       \iffieldundef{labelyear}
         {}
         {\printfield{labelyear}%
          \printfield{extrayear}}}}
    {\usebibmacro{cite:shorthand}}}

\AtEveryBibitem{\clearfield{month}\clearfield{note}}

\usepackage{hyperref}
\definecolor{darkforestgreen}{RGB}{30, 96, 32}
\definecolor{darknavy}{RGB}{0, 64, 128}
\hypersetup{
    colorlinks=true,
    linkcolor=darknavy,        
    citecolor=darkforestgreen, 
    urlcolor=darknavy,         
}

\usepackage{amsmath,
amssymb,
amsthm,
mathtools}

\usepackage{tikz-cd}
\usetikzlibrary{positioning,arrows.meta,calc,fit,backgrounds}

\newcommand{\kyle}[1]{}
\newcommand{\nate}[1]{}

\usepackage[capitalize,noabbrev]{cleveref}

\theoremstyle{plain}
\newtheorem{theorem}{Theorem}[section]

\theoremstyle{definition}
\newtheorem{definition}[theorem]{Definition}

\theoremstyle{remark}

\theoremstyle{definition}
\newtheorem{example}[theorem]{Example}

\newcommand\cO{\mathcal{O}}

\newcommand\cQ{\mathcal{Q}}

\newcommand\cV{\mathcal{V}}

\usepackage[disable,textsize=tiny]{todonotes}
\usepackage{fontawesome5}

\title{Operadic consistency: a label-free signal for compositional reasoning failures in LLMs}

\author{%
  Nathaniel Bottman \\
  Incubilate \\
  Seattle, USA \\
  \href{mailto:nate@incubilate.com}{\texttt{nate@incubilate.com}} \\
  \And
  Yinhong Liu \\
  University of Cambridge \\
  Cambridge, UK \\
  \href{mailto:yl535@cam.ac.uk}{\texttt{yl535@cam.ac.uk}}
  \And
  Kyle Richardson \\
  Allen Institute for Artificial Intelligence \\
  Seattle, USA \\
  \href{mailto:kyler@allenai.org}{\texttt{kyler@allenai.org}}
}

\begin{document}

\maketitle

\begin{center}
  \faGithub\, \href{https://github.com/natebottman/operadic-consistency-paper}{\texttt{github.com/natebottman/operadic-consistency-paper}}
\end{center}

\begin{abstract}
Detecting LLM reasoning failures at inference time without ground-truth labels has motivated a wide range of confidence baselines, including self-consistency, semantic entropy, and P(True), built on within-question sampling and self-evaluation. Operad theory, the formalism for systems built by iterated substitution, suggests a complementary diagnostic: a model's direct answer to a compositional query should agree with the answer it produces by composing a stated decomposition of the same query. We instantiate this idea as operadic consistency (OC), a per-question signal. Across twelve instruction-tuned LLMs (4B to 671B parameters, open-weights and closed-source) on four multi-hop QA datasets, OC is strongly correlated with accuracy on every dataset (Pearson $r \in [0.86, 0.94]$, all $p \leq 0.0004$), and is the only signal we evaluate with $r \geq 0.85$ uniformly across all four datasets. Chain-of-thought self-consistency (CoT-SC; Wang et al., 2023) matches OC on HotpotQA and DROP ($r = 0.93, 0.87$) but drops to $r \approx 0.45$ on MuSiQue and StrategyQA. At the per-question level, OC contributes information beyond CoT-SC and semantic entropy on every dataset (cluster-robust $p \leq 10^{-16}$ for the OC coefficient), and the conclusion is robust to additionally controlling for constructed decomposition-aware baselines ($p \leq 10^{-13}$). The same signal yields selective-prediction improvements (accuracy at fixed coverage) over a tuned CoT-SC baseline at the equal-cost $K = 3$ budget (AUARC lifts of +0.086 to +0.096 and AUROC lifts of +0.092 to +0.164; 95\% CIs exclude zero on every cell). On five frontier thinking models, where the decomposition is extracted from the model's own chain of thought, the same equal-cost comparison gives positive selective-prediction point-estimate lift on all 16 (dataset, budget, metric) cells tested, with 95\% CIs excluding zero on 12 of the 16.
\end{abstract}

\section{Introduction}
\label{s:intro}

Modern LLMs (including the latest generation of explicit ``thinking'' models) routinely produce extended chains of reasoning that look fluent and confident but do not actually compose into a correct answer \citep{press2022compositionality, lanham2023measuring}.
Detecting these failures at inference time, without ground-truth labels, is the practical question behind work on hallucination detection, calibration, and selective prediction.
Existing methods cluster into a few families: \emph{self-consistency} \citep{wang2022self} aggregates $K$ temperature samples by majority vote; as a continuous confidence score we use mean pairwise agreement across the same samples. \emph{Semantic entropy} \citep{kuhn2023semantic} computes Shannon entropy over semantic clusters of samples. \emph{$P(\text{True})$} \citep{kadavath2022language} elicits the model's own probability that its answer is correct.
The first two probe sample diversity (``how variable is the model's answer when re-sampled''); the third probes self-evaluation (``does the model think it is right'').
A complementary line of work \citep{jacovi-etal-2024-chain, akbar-etal-2024-hallumeasure} employs external verification, using fact-checking tools or process reward models to evaluate decomposed atomic claims. They typically depend on task-specific tooling or partial supervision.

\begin{figure}[t]
\centering
\includegraphics[width=0.75\linewidth]{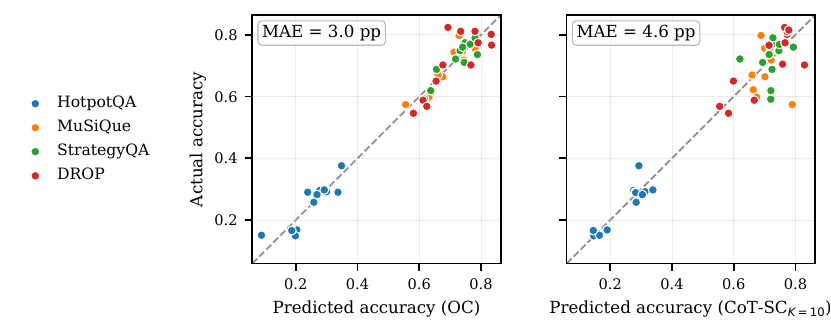}
\caption{\textbf{Operadic consistency yields a label-free cross-model accuracy estimator.} For each non-thinking model and dataset, we hold the model out, fit \(\text{accuracy} \approx a\cdot s + b\) on the remaining eleven models for a per-(model, dataset) signal-rate \(s\), and predict the held-out model's accuracy. \emph{Left:} \(s = \text{OC-rate}\) gives pooled MAE \(3.0\) pp (range $2.1$--$4.7$ pp per dataset) with no calibration set or ground-truth labels for the held-out model. \emph{Right:} \(s = \text{CoT-SC}_{K{=}10}\)-rate (canonical chain-of-thought self-consistency, \citealp{wang2022self}); ten inference calls per question rather than three. Pooled MAE \(4.6\) pp, but with a bimodal per-dataset pattern: CoT-SC matches OC on HotpotQA and DROP (per-dataset MAE $2.4$ and $4.9$ pp) and loses uniform predictive value on MuSiQue and StrategyQA ($5.8$ and $5.2$ pp).}
\label{fig:teaser}
\end{figure}
None asks whether the model's reasoning is \emph{internally compositional} --- whether the sub-conclusions, taken seriously, would actually compose into the same final answer the model gives directly.

We make this latter notion precise.
Many of the reasoning tasks LLMs are evaluated on (multi-hop question answering, math word problems, multi-step planning) have an explicit \emph{compositional} structure: a complex query decomposes into simpler sub-queries whose answers are substituted to produce a final answer. \citet{wei2022chain} and a long line of subsequent work \citep{yao2023tree, khot2022decomposed, yang2024buffer, besta2024graph} use this structure operationally, prompting models to expose intermediate reasoning steps.
We use it diagnostically: when a model gives a direct answer to a complex query and \emph{also} operates on a stated decomposition of that query (whether supplied externally or extracted from the model's own chain of thought), the two answers should agree if the model's reasoning is internally consistent.
This failure mode is the per-question analog of the population-level \emph{compositionality gap} of \citet{press2022compositionality}, who show that LLMs routinely fail to compose correct sub-answers into correct multi-hop answers.
We call this check \emph{operadic consistency}, and we show it is a strong, complementary signal of per-question correctness.

We propose that the mathematical framework underlying this signal is the formalism of \emph{operads} \citep{markl_stasheff_shnider:operads_in_algebra_topology_and_physics, loday_vallette:algebraic_operads}, a well-studied structure introduced to study systems built by iterated substitution --- how many-input operations compose coherently into larger operations --- and providing a compact language for tree-shaped composition.
We define the \emph{questions operad} $\cQ$, in which operations correspond to question templates with blanks and composites correspond to decompositions into sub-queries, and interpret a language model as an algebra over $\cQ$.
The operad framework is not just descriptive vocabulary: it is what \emph{suggests} the notion of operadic consistency --- the predicate that direct and decomposed answers agree on a given tree of questions.
The framework itself (general definitions, examples from formal language theory, and operadic consistency on arbitrary trees of questions) is developed in our companion paper \citep{bottman:operads_workshop}; the present paper uses only its depth-2 instantiation.

We instantiate the framework on five compositional reasoning benchmarks (HotpotQA, MuSiQue, StrategyQA, DROP, GSM8K), across a cohort of seventeen models: twelve instruction-tuned LLMs (4B--671B parameters, spanning open-weights and closed-source frontier offerings) and five frontier thinking models.

\paragraph{Contributions.} In summary:
\begin{enumerate}
  \item \emph{Framework.} We define the questions operad $\cQ$ and interpret a language model as an algebra over $\cQ$. From this we derive \emph{operadic consistency} (the predicate that direct and decomposed answers agree on every tree of questions), and use its scalar instantiation OC as a per-question signal (\S\ref{s:operads}).
  \item \emph{Empirical signal.} Across twelve instruction-tuned LLMs (4B--671B) on four multi-hop QA datasets, OC correlates strongly with cross-model accuracy ($r \in [0.86, 0.94]$) and is the only signal we evaluate with $r \geq 0.85$ on all four datasets; canonical chain-of-thought self-consistency \citep{wang2022self} matches OC on HotpotQA and DROP but drops to $r \approx {+}0.45$ on MuSiQue and StrategyQA. Under a leave-one-model-out fit, the OC-rate of a held-out model predicts its accuracy to within $3.0$ pp on average, well below the $4.6$ pp pooled obtained from the same procedure with CoT-SC at $K{=}10$ (Fig.~\ref{fig:teaser}). At the per-question level, OC contributes information beyond CoT-SC, semantic entropy, and constructed decomposition-aware baselines on every dataset (\S\ref{ss:across_model}).
  \item \emph{Deployment lifts.} In a leave-one-model-out protocol (the calibrator is fit on eleven models and evaluated on the held-out twelfth, simulating deployment to a new model), OC combined with $\text{CoT-SC}_{K=3}$ improves on selective-prediction metrics over a tuned $\text{CoT-SC}_{K=3}$ baseline (the equal-cost comparator at $3$ model calls) by $\Delta_{\text{AUARC}} \in [{+}0.086, {+}0.096]$ and $\Delta_{\text{AUROC}} \in [{+}0.092, {+}0.164]$ on the four datasets, with bootstrap $95\%$ CIs excluding zero on every cell (\S\ref{ss:across_model}).
  \item \emph{Thinking models.} On five frontier thinking models, with the decomposition extracted from the model's own chain of thought, the same equal-cost comparison gives positive selective-prediction lift on all $16$ (dataset, budget, metric) cells tested (\S\ref{ss:thinking_models}).
\end{enumerate}

\section{Operads}
\label{s:operads}

We use the language of \emph{operads} to make the compositional structure of question decomposition precise.
An operad organizes operations of varying arities ($k$ inputs and one output) together with composition operations $\circ_i\colon \cO(k) \times \cO(\ell) \to \cO(k+\ell-1)$ that plug one operation's output into the $i$-th input slot of another, satisfying an \emph{associativity axiom}.
Formal definitions are in our companion paper \citep{bottman:operads_workshop}; for accessible references see \citet{markl_stasheff_shnider:operads_in_algebra_topology_and_physics, loday_vallette:algebraic_operads}.

\subsection{The questions operad}
\label{ss:operads_for_QD}

The questions operad sits within the space of decomposed-prompting frameworks: chain-of-thought \citep{wei2022chain} and its variants \citep{khot2022decomposed, hu2024uncertainty, yao2023tree} all surface intermediate sub-questions, and the questions operad organizes the resulting set of templates and their compositions into a single algebraic object.
The questions operad organizes \emph{question templates} (natural-language questions with blanks) and their compositions.
Crucially, composition is purely an operation on question templates, separate from the algebra step where a model actually produces answers; we make both notions precise below.

\begin{definition}[The questions operad]
\label{def:Q_sketch}
The (non-unital) \emph{questions operad $\cQ$} has $\cQ(k)$ the set of question templates with $k$ blanks. For every $k, \ell \geq 0$ and $1 \leq i \leq k$, the composition operation $\circ_i\colon \cQ(k) \times \cQ(\ell) \to \cQ(k + \ell - 1)$ is defined by \emph{insertion of question templates}: the second template (with its $\ell$ blanks) is substituted in place of blank $i$ of the first, yielding a new template with $k+\ell-1$ blanks. The composition operations satisfy associativity in the standard operadic sense (\citealp{bottman:operads_workshop}); on natural-language renderings, associativity holds up to semantic equivalence (the linguistic surface form of a composite may vary, but the meaning is the same).
\end{definition}

\begin{example}
\label{ex:WW2}
For a concrete instance, let $\mathtt{Q1}, \mathtt{Q2}, \mathtt{Q3} \in \cQ$ be the templates ``When did World War 2 end?'', ``Who was President at $-$?'', and ``Who was $-$'s wife?'' respectively, with $\mathtt{Q1} \in \cQ(0)$ and $\mathtt{Q2}, \mathtt{Q3} \in \cQ(1)$. Their composite $\mathtt{Q3}\circ_1\mathtt{Q2}\circ_1\mathtt{Q1} \in \cQ(0)$ is the question ``Who was First Lady when World War 2 ended?''.
\end{example}

\paragraph{Algebras over $\cQ$, and language models.}
An \emph{algebra} over $\cQ$ interprets each template as a concrete function on values: fix a value set $\cV$, and assign to each $q \in \cQ(k)$ a map $\cV(q)\colon \cV^k \to \cV$, compatibly with composition (\citealp{bottman:operads_workshop}).
A language model $m$ determines a candidate algebra in the natural way: substitute the values into $q$'s blanks, then ask the model.
The associativity axiom would say that $\cV_m$ is in fact a $\cQ$-algebra: the model's answer to a composite question agrees with what one obtains by sequentially answering its parts.
Its failure is what we measure as \emph{operadic consistency} (\S\ref{s:empirics}).

\subsection{Operadic consistency}
\label{ss:operadic_consistency}
A \emph{tree of questions} (ToQ) $T$ is a rooted, oriented, planar tree where edges point toward the root: each arity-$k$ vertex has $k$ incoming edges (one per question blank) and one outgoing edge (the answer it produces). Leaves are arity-$0$ vertices with no incoming edges; the root has a distinguished outgoing edge representing the final answer of the composite. Each arity-$k$ vertex carries a question $q \in \cQ(k)$.

The \emph{total composite} of $T$ is the question obtained by composing along every internal edge.
Given a model $m$, two answers to the total composite are available:
the \emph{direct answer}, asking $m$ the total composite as a single question; and
the \emph{decomposed answer}, evaluating $T$ leaf-to-root: ask $m$ each leaf question, then for each non-leaf vertex substitute its children's answers into its blanks and ask $m$.

\begin{definition}
\label{def:operadic_consistency}
$m$ is \emph{operadically consistent on $T$} if its direct and decomposed answers on $T$ agree.
\end{definition}

The empirics of \S\ref{s:empirics} evaluate operadic consistency on the simplest non-trivial ToQ, a depth-2 chain (one internal edge; Fig.~\ref{fig:oc_pipeline}), and replace the binary predicate of Def.~\ref{def:operadic_consistency} with a continuous score in $[0,1]$ measuring agreement under a per-dataset semantic equivalence, the \emph{OC score}.
The choice of equivalence is itself constrained by the framework: the OC scorer should realize the natural equivalence on the answer space appropriate to the question's expected output type. App.~\ref{app:drop_sensitivity} tests this empirically.

\section{Empirical evidence}
\label{s:empirics}

We present two empirical findings:
\S\ref{ss:across_model} shows that across non-thinking models, operadic consistency is strongly correlated with accuracy where temperature self-consistency is not, and contributes predictive information about per-question correctness beyond standard sample-based, self-evaluation, and decomposition-aware baselines.
\S\ref{ss:thinking_models} extends the analysis to thinking models, using operadic structure extracted from a model's own chain of thought.

\subsection{Operadic consistency as a per-question selection signal}
\label{ss:across_model}

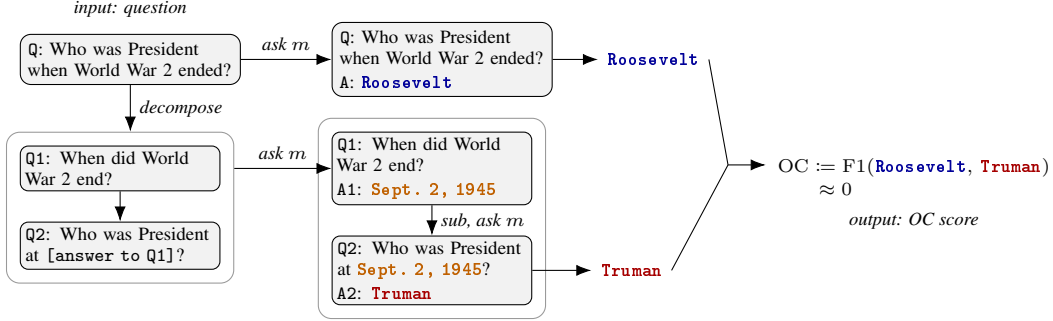
\begin{figure}[t]
\centering
\begin{tikzpicture}[
  >={Latex[length=2mm]},
  qbox/.style={
    draw, rounded corners=4pt, align=left,
    fill=black!5, inner sep=3pt, font=\scriptsize,
    text width=2.7cm
  },
  qinner/.style={
    draw, rounded corners=3pt, align=left,
    fill=black!5, inner sep=2pt, font=\scriptsize,
    text width=2.5cm
  },
  outer/.style={
    draw=black!40, rounded corners=5pt, inner sep=5pt
  },
  c1/.style={blue!60!black},     
  c2/.style={orange!75!black},   
  c3/.style={red!65!black}       
]
\node[qbox] (q) {\texttt{Q}: Who was President when World War 2 ended?};
\node[font=\scriptsize\itshape, anchor=south] at ([yshift=1mm]q.north)
  {input: question};
\node[qbox, right=12mm of q] (qans)
  {\texttt{Q}: Who was President when World War 2 ended?\\[1pt]
   \texttt{A}: {\color{blue!60!black}\textbf{\texttt{Roosevelt}}}};
\draw[->] (q.east) -- node[above, font=\scriptsize\itshape] {ask $m$} (qans.west);

\node[qinner, below=8mm of q.south west, anchor=north west] (q1)
  {\texttt{Q1}: When did World War 2 end?};
\node[qinner, below=4mm of q1] (q2)
  {\texttt{Q2}: Who was President at \texttt{[answer to Q1]}?};
\draw[->] (q1.south) -- (q2.north);
\begin{scope}[on background layer]
\node[outer, fit=(q1)(q2)] (coq) {};
\end{scope}

\draw[->] (q.south) -- node[right, font=\scriptsize\itshape] {decompose} (q.south |- coq.north);

\node[qinner, anchor=west] (a1) at (qans.west |- q1)
  {\texttt{Q1}: When did World War 2 end?\\[1pt]
   \texttt{A1}: {\color{orange!75!black}\textbf{\texttt{Sept.\ 2, 1945}}}};
\node[qinner, below=4mm of a1] (a2)
  {\texttt{Q2}: Who was President at {\color{orange!75!black}\textbf{\texttt{Sept.\ 2, 1945}}}?\\[1pt]
   \texttt{A2}: {\color{red!65!black}\textbf{\texttt{Truman}}}};
\draw[->] (a1.south) -- node[right, font=\scriptsize\itshape] {sub, ask $m$} (a2.north);
\begin{scope}[on background layer]
\node[outer, fit=(a1)(a2)] (chain_ans) {};
\end{scope}

\draw[->] (coq.east |- q1) -- node[above, font=\scriptsize\itshape] {ask $m$} (a1.west);

\node[font=\scriptsize, anchor=west] (adirect) at ([xshift=6mm]qans.east)
  {{\color{blue!60!black}\textbf{\texttt{Roosevelt}}}};
\draw[->] (qans.east) -- (adirect.west);

\node[font=\scriptsize, anchor=west] (a2out) at ([xshift=6mm]chain_ans.east |- a2)
  {{\color{red!65!black}\textbf{\texttt{Truman}}}};
\draw[->] (a2.east) -- (a2out.west);

\coordinate (meet) at ($(adirect.east)!0.5!(a2out.east) + (5mm,0)$);
\coordinate (tip)  at ([xshift=5mm]meet);
\draw (adirect.east) -- (meet);
\draw (a2out.east)   -- (meet);
\draw[->] (meet) -- (tip);
\node[anchor=west, font=\scriptsize, inner sep=0pt] (oc) at ([xshift=1.5mm,yshift=-0.3mm]tip)
  {$\mathrm{OC} \coloneqq \mathrm{F1}(\textcolor{blue!60!black}{\textbf{\texttt{Roosevelt}}},\,\textcolor{red!65!black}{\textbf{\texttt{Truman}}})$};
\node[anchor=base west, font=\scriptsize, inner sep=0pt] (ocb) at ([yshift=-3mm]oc.base west)
  {$\phantom{\mathrm{OC}}\,\approx 0$};
\node[font=\scriptsize\itshape, anchor=north] at ([yshift=-1mm]ocb.south -| oc)
  {output: OC score};
\end{tikzpicture}
\caption{Operadic-consistency (OC) check on a depth-2 chain (Ex.~\ref{ex:WW2}), where $m$ denotes the language model being evaluated.
The undecomposed question $\mathtt{Q}$ (top-left) decomposes into the chain $(\mathtt{Q1}, \mathtt{Q2})$ (bottom-left). The model is asked two ways: directly on $\mathtt{Q}$ (top), and step-by-step on the chain (bottom) by answering $\mathtt{Q1}$, substituting the result into $\mathtt{Q2}$, and answering again. Model outputs are color-coded: the direct answer to $\mathtt{Q}$ in blue, the answer to $\mathtt{Q1}$ (which carries through the substitution) in orange, and the answer to $\mathtt{Q2}$ in red. OC compares the direct and decomposed final answers under the per-dataset \texttt{score\_official} scorer, which we abbreviate here as F1; on this example OC $\approx 0$, flagging the question for selective prediction.}
\label{fig:oc_pipeline}
\end{figure}

We first study operadic consistency on instruction-tuned (non-thinking) models, where the decomposition tree is supplied by human annotators rather than extracted from the model itself. Figure~\ref{fig:oc_pipeline} illustrates the OC computation on the depth-2 chain from Ex.~\ref{ex:WW2}.
We compare against three standard uncertainty-quantification baselines: \emph{canonical chain-of-thought self-consistency} (abbreviated $\text{CoT-SC}$) \citep{wang2022self}, \emph{semantic entropy} ($H$) \citep{kuhn2023semantic}, and the Kadavath self-evaluation baseline $P(\text{True})$ \citep{kadavath2022language}.
We use $\text{OC}$ as shorthand for our \emph{operadic consistency} signal.

\paragraph{Setup.}
We evaluate twelve instruction-tuned models spanning 4B to 671B parameters (Llama-3 8B, Gemma-3n 4B, Qwen2.5 7B, LiquidAI 24B, Llama-3.3 70B, Qwen3-235B, DeepSeek-V3.1, Cogito 671B, GPT-4o-mini, Gemini 2.5 Flash, Llama-4 Maverick, Mistral Large 2) on four multi-hop QA datasets (HotpotQA, MuSiQue, StrategyQA, DROP; 403--625 questions each), using two-step decompositions from Break QDMR annotations \citep{wolfson2020break} for HotpotQA and DROP, and native decompositions for MuSiQue and StrategyQA.\footnote{GSM8K lacks Break-style two-step decomposition annotations and enters only in \S\ref{ss:thinking_models}, where decompositions come from thinking traces.}
For each (model, question) pair we compute:
\begin{itemize}
\item \emph{Operadic consistency} ($\text{OC}$): semantic-agreement score between the model's greedy direct answer and its greedy decomposed answer (answer step~1, substitute into step~2, answer). Per-dataset \texttt{score\_official} agreement (SQuAD-F1 for HotpotQA/MuSiQue, yes/no extraction for StrategyQA, value-equality for DROP; full scorer in App.~\ref{app:stats}). The score lies in $[0,1]$, used as-is in regressions and thresholded at $0.5$ when a binary partition is needed.
Cost: 3 model calls.
\item \emph{Canonical CoT self-consistency} follows the $K$-sample CoT protocol of \citet{wang2022self}: sample $K$ chain-of-thought responses at temperature $0.7$ (zero-shot CoT prompt; \citealp{kojima2022large}) and extract a short answer from each chain via a deterministic auxiliary LLM extractor (Llama-3.3-70B at $T{=}0$; details in App.~\ref{app:baselines}). As a continuous confidence score we use the mean pairwise per-dataset \texttt{score\_official} agreement over the $K$ extracted answers, for $K{=}3$ and $K{=}10$. Pairwise comparisons use the per-dataset accuracy scorer (yes/no extraction for StrategyQA, value-equality for DROP, SQuAD-F1 elsewhere) so that CoT-SC and OC are scored on the same equivalence; see App.~\ref{app:drop_sensitivity}.
Cost: $K$ chain-generation calls plus $K$ cheap extractor calls.
\item \emph{Semantic entropy} \citep{kuhn2023semantic}: Shannon entropy over clusters of the $K{=}10$ samples, where two samples cluster together if their matched-axis pairwise agreement is $\geq 0.5$ (App.~\ref{app:drop_sensitivity}).
Cost: $K{=}10$ calls plus clustering.
\item \emph{$P(\text{True})$} \citep{kadavath2022language}: ask the model ``Question: \textit{q}.
Proposed answer: \textit{a}.
Is the proposed answer correct? Reply Yes or No.'' and read $P(\text{Yes}) / (P(\text{Yes}) + P(\text{No}))$ from the next-token logprobs.
Cost: 1 model call.
\end{itemize}
\emph{Accuracy} for each (model, question) pair is the indicator that the greedy direct answer agrees with the gold answer at per-dataset \texttt{score\_official} $\geq 0.5$.
We report per-question regressions at $K{=}10$ (the literature-default budget at which CoT-SC and $-H$ are at full strength) and additionally include the equal-cost $K{=}3$ budget in selective-prediction comparisons (matching OC's $3$ model calls).

\begin{figure*}[!t]
  \centering
  \includegraphics[width=\textwidth]{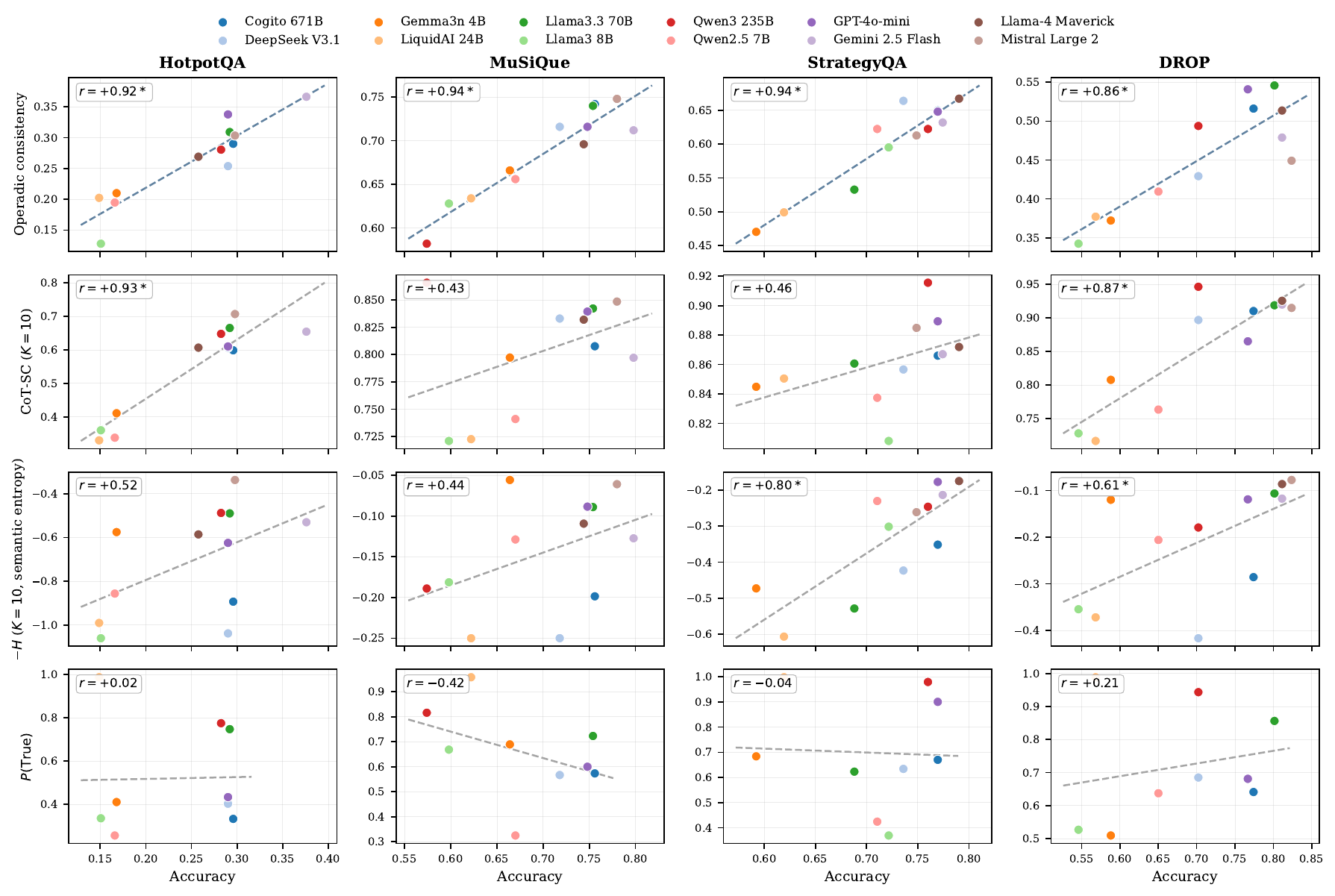}
  \caption{Signal vs.\ accuracy across twelve models and four datasets, with one row per uncertainty-baseline family.
  \textbf{Row 1:} operadic consistency, the proposal ($r \in [0.86, 0.94]$, all $p \leq 0.0004$).
  \textbf{Row 2:} canonical CoT self-consistency at $K{=}10$ \citep{wang2022self}, the dominant sample-aggregate baseline. Bimodal: $r{=}{+}0.93, {+}0.43, {+}0.46, {+}0.87$ across the four datasets.
  \textbf{Row 3:} semantic entropy at $K{=}10$ \citep{kuhn2023semantic}.
  \textbf{Row 4:} $P(\text{True})$ self-evaluation \citep{kadavath2022language}.
  OC is the only row with $r \geq 0.85$ on every dataset; every other row has at least one dataset where $r < 0.85$. Full per-baseline table (including naive direct-answer SC, decomposed-SC, and PRMs) in App.~\ref{app:cross_model_baselines}.
  Each point is one model; trend line and Pearson $r$ with two-sided $p$ shown per subplot ($*$ marks $p<0.05$). HotpotQA is evaluated closed-book (no passage), hence the lower per-model accuracies; see App.~\ref{app:datasets}.}
  \label{fig:empirics}
\end{figure*}

\paragraph{Across-model finding.}
Figure~\ref{fig:empirics} plots each consistency metric against accuracy across the twelve models, per dataset.
Operadic consistency is strongly correlated with accuracy on every dataset (Pearson $r$ between $0.86$ and $0.94$; $p \leq 0.0004$ on all four).
Canonical CoT-SC shows a \emph{bimodal} pattern: on HotpotQA and DROP it matches OC ($r{=}{+}0.93$ and ${+}0.87$, both $p < 0.001$), but on MuSiQue and StrategyQA it collapses to $r{=}{+}0.43$ and ${+}0.46$ (both n.s.). No other sample-based or self-evaluation baseline we test achieves $r \geq 0.85$ on more than two datasets: semantic entropy $-H_{K=10}$ peaks at $+0.80$ on StrategyQA but drops to $+0.44$ on MuSiQue; $P(\text{True})$ shows no significant correlation on any dataset (App.~\ref{app:cross_model_baselines}).
OC remains the only baseline with $r \geq 0.85$ uniformly across all four datasets, consistent with the broader observation that sample-based methods probe within-question confidence (largely orthogonal to cross-model competence: a confidently-wrong model has low sample variance just as a confidently-correct one does), while OC probes a complementary compositional axis.
This relationship is robust to the choice of correctness scorer: under an LLM-judge correctness scorer (Claude Haiku 4.5 with a paraphrase-tolerant prompt), the same cross-model correlation gives $r \in [{+}0.69, {+}0.97]$, all $p \leq 0.013$ (App.~\ref{app:drop_sensitivity}).
Practically, the strong cross-model correlation immediately yields a label-free dataset-level \emph{accuracy estimator}: a leave-one-model-out demonstration on the twelve non-thinking models (App.~\ref{app:dataset_pred}) achieves mean absolute prediction error of $3.0$ pp pooled (range $2.1$--$4.7$ pp across datasets), relevant when ground-truth labels are scarce.

\paragraph{Per-question complementarity.}
Within a single model, do operadically-consistent questions get answered correctly more often, even controlling for canonical CoT-SC and semantic entropy? On the same twelve models pooled across the four datasets ($n=24{,}624$ (model, question) pairs), a per-dataset logistic regression of correctness on the three continuous signals $(\text{OC}, \text{CoT-SC}_{K=10}, -H_{K=10})$ gives the coefficients in Table~\ref{tab:nonthinking_combined}; standard errors are cluster-robust, clustered by question. All three predictors are highly significant on all four datasets ($p \leq 10^{-16}$ for $\beta_{\text{OC}}$ on every dataset; $p \leq 10^{-16}$ for $\beta_{\text{CoT-SC}}$ on every dataset); the three signals are clearly complementary, no single one subsumes another.

\paragraph{Selective prediction.}
The regression coefficients translate into gains at roughly one third the inference cost of the strongest sample-based baseline ($3$ vs.\ $10$ model calls).
For each $(\text{model}, \text{dataset})$ cell we compute AUARC \citep{geifman2017selective} and AUROC of a logistic-regression combiner taking $(\text{OC}, \text{CoT-SC}_{K=3})$, trained on the other eleven models pooled and evaluated on the held-out cell (leave-one-model-out, so the calibrator is never fit on the cell it scores).
At the equal-cost $K{=}3$ budget, OC$+$CoT-SC improves AUARC by $+0.086$ to $+0.096$ and AUROC by $+0.092$ to $+0.164$ over $\text{CoT-SC}_{K=3}$ alone (right panel of Table~\ref{tab:nonthinking_combined}); bootstrap $95\%$ CIs over the twelve paired model-level differences exclude zero on every dataset.
Lifts against the literature-default $\text{CoT-SC}_{K=10}$ budget are uniformly smaller but still positive with CIs excluding zero on every dataset (App.~\ref{app:selective_prediction}).

\begin{table}[h]
\centering
\small
\begin{minipage}{0.55\linewidth}
\centering
\setlength{\tabcolsep}{3pt}
\begin{tabular}{l rrr}
\toprule
dataset & $\beta_{\text{OC}}\ (p)$ & $\beta_{\text{CoT-SC}}\ (p)$ & $\beta_{-H}\ (p)$ \\
\midrule
HotpotQA   & $+1.90\ (10^{\text{-}38})$ & $+1.86\ (10^{\text{-}27})$ & $+0.94\ (10^{\text{-}30})$ \\
MuSiQue    & $+2.24\ (10^{\text{-}43})$ & $+1.88\ (10^{\text{-}16})$ & $+1.37\ (10^{\text{-}25})$ \\
StrategyQA & $+1.69\ (10^{\text{-}37})$ & $+1.93\ (10^{\text{-}19})$ & $+1.29\ (10^{\text{-}82})$ \\
DROP       & $+1.19\ (10^{\text{-}16})$ & $+2.24\ (10^{\text{-}23})$ & $+2.09\ (10^{\text{-}53})$ \\
\bottomrule
\end{tabular}
\end{minipage}\hfill
\begin{minipage}{0.42\linewidth}
\centering
\setlength{\tabcolsep}{4pt}
\begin{tabular}{l rr}
\toprule
dataset & $\Delta_{\text{AUARC}}$ & $\Delta_{\text{AUROC}}$ \\
\midrule
HotpotQA   & $+0.095$ & $+0.092$ \\
MuSiQue    & $+0.086$ & $+0.153$ \\
StrategyQA & $+0.096$ & $+0.164$ \\
DROP       & $+0.086$ & $+0.117$ \\
\bottomrule
\end{tabular}
\end{minipage}
\\[6pt]
\caption{\emph{Left:} three-way logistic regression of correctness on $(\text{OC}, \text{CoT-SC}_{K=10}, -H_{K=10})$ (all continuous), fit per dataset on twelve non-thinking models, with cluster-robust standard errors clustered by question. All three predictors are massively significant ($p \leq 10^{-16}$) on every dataset, indicating that OC, CoT-SC, and semantic entropy each carry complementary marginal information; no single signal subsumes the others. App.~\ref{app:p_true_sensitivity} reports a four-way variant adding $P(\text{True})$.
\emph{Right:} selective-prediction lift from adding OC to a logistic-regression combiner over $\text{CoT-SC}_{K=3}$, leave-one-model-out across the same twelve models; each cell is the paired mean of [AUARC/AUROC of OC$+$CoT-SC] minus [AUARC/AUROC of CoT-SC alone] at the equal-cost ($K{=}3$) budget. Bootstrap $95\%$ CIs (full intervals in App.~\ref{app:selective_prediction}) exclude zero on every cell. Lifts vs the literature-default $\text{CoT-SC}_{K=10}$ are uniformly smaller but still all positive with CIs excluding zero (App.~\ref{app:selective_prediction}).}
\label{tab:nonthinking_combined}
\end{table}

\paragraph{OC outperforms decomposition-aware baselines.}
The baselines compared so far (CoT-SC, semantic entropy, and $P(\text{True})$) share a common feature: none uses the question decomposition.
This raises two concrete concerns about what OC's positive correlation with accuracy is actually measuring.
\emph{Concern (i): OC is self-consistency in disguise.} The decomposed-path answer is just one model output, and if the model is unstable on the decomposed path, OC's agreement with the direct answer might track per-question variance on that path rather than any compositional property; the same question signal would be obtainable from $K$-sample self-consistency restricted to the decomposed-path answer.
\emph{Concern (ii): OC is a worse process reward model.} An off-the-shelf process reward model (PRM) scored on the decomposed steps could plausibly capture everything OC captures and more, in which case OC adds no information beyond cheap-PRM features.
We test both with constructed decomposition-aware baselines (each combines standard machinery with our specific decomposition protocol; neither is a published off-the-shelf comparison):
(i) \emph{decomposed self-consistency} (our continuous pairwise-agreement adaptation of self-consistency \citep{wang2022self}, applied to $K{=}10$ samples of the decomposed-path answer; see App.~\ref{app:baselines});
and (ii) \emph{Skywork-PRM}, an open process reward model \citep{he2024skyworko1, lightman2024lets} trained on Skywork-o1 step preferences for math and code reasoning, applied off-label to the $(\mathtt{Q1}, \mathtt{a1}, \mathtt{Q2}, \mathtt{a2})$ trace.

At the cross-model level, both decomposition-aware baselines show isolated significant correlations on a single dataset each: decomposed-SC on HotpotQA ($r{=}{+}0.67$, $p{=}0.018$) and Skywork-PRM on DROP ($r{=}{+}0.82$, $p{=}0.001$). Neither correlates significantly on the other three datasets, leaving OC as the only baseline with $r \geq 0.85$ uniformly. At the per-question level, we re-run the logistic regression with each new baseline added as a \emph{fourth} predictor alongside OC, $\text{CoT-SC}_{K=10}$, and $-H_{K=10}$, fit per dataset on the same twelve non-thinking models with cluster-robust standard errors clustered by question.
Across all four datasets the OC coefficient remains massively significant ($\beta_{\text{OC}} \geq +1.11$, $p \leq 10^{-13}$; full coefficients in App.~\ref{app:dec_aware_app}).
Decomposed self-consistency's coefficient is non-significant or \emph{negative} on every dataset (significantly negative on HotpotQA and StrategyQA), decisively rejecting (i).
Skywork-PRM contributes independent information on two of four datasets but never displaces OC, addressing (ii) modulo the off-label use.

\subsection{Operadic consistency for thinking models}
\label{ss:thinking_models}

The decomposition chains in \S\ref{ss:across_model} are supplied by human annotators (Break QDMR, MuSiQue's native decompositions, etc.).
For modern thinking models, an alternative is available: the model's own chain of thought already exhibits a decomposition into intermediate sub-questions and sub-answers.
We use this to extend the operadic consistency check to thinking models without requiring external decomposition annotations. The reliability of thinking-model confidence is itself an active question: \citet{mei2025reasoning} report systematic overconfidence on incorrect responses; \citet{yoon2025reasoning} find verbalized confidence better-calibrated in reasoning models than in their non-reasoning siblings on most settings; \citet{podolak2025readmind} show that these gains accrue from additional reasoning prior to confidence elicitation, with the resulting verbalized confidence largely recoverable from the trace alone by an auxiliary model; and on the abstention side, \citet{kirichenko2025abstentionbench} document a $24$-point average drop in abstention quality from reasoning fine-tuning, across 20 unanswerable-question datasets. OC supplies a complementary, decomposition-based signal that does not depend on either sampling diversity or verbalized self-assessment.

\paragraph{CoQ extraction from CoT.}
For each (model, question) pair we generate one greedy thinking trace and apply a lightweight extractor (an instruction-tuned model prompted to identify the first explicit sub-question and the model's answer to it) to lift a two-step chain $\mathtt{Q1} \to \mathtt{Q2}$ from the trace.
We discard pairs where the extractor flags the trace as not factorable (no clean sub-question structure).
The resulting depth-2 ToQ is the model's own implicit decomposition; we then run the operadic-consistency check exactly as in \S\ref{ss:across_model}: compare the model's direct greedy answer to the answer it gives when asked to substitute $\mathtt{[A1]}$ into $\mathtt{Q2}$ and re-answer.

\paragraph{Setup.}
We evaluate five thinking models (DeepSeek-R1, GLM-5.1, Cogito v2.1, Kimi-K2.6, MiniMax-M2.7) on four multi-hop QA datasets (MuSiQue, StrategyQA, GSM8K, DROP), $100$ questions per (model, dataset) cell at one seed.
After removing extractor-flagged unfactorable traces and small per-cell losses to API errors, we retain roughly $94$ questions per cell on average.
$P(\text{True})$ is not available for thinking endpoints in our protocol (no top-$k$ logprobs after the $\langle\text{think}\rangle$ tag), so the Kadavath baseline is omitted here.

Because thinking models produce reasoning chains natively, the self-consistency variant compared throughout this section is structurally identical to the canonical CoT-SC of \S\ref{ss:across_model}: $K$ samples at $T{=}0.7$, the same Llama-3.3-70B short-answer extractor (App.~\ref{app:baselines}), and matched-axis pairwise per-dataset \texttt{score\_official} agreement. For compactness, we abbreviate it $\text{SC}$ here.

\paragraph{Equal-cost comparison.}
Operadic consistency on a depth-2 ToQ costs $3$ thinking calls per question (direct, sub-question 1, sub-question 2). $\text{SC}_{K=3}$ also costs $3$ thinking calls; $-H_{K=3}$ uses the same three samples and is essentially free additional information on top of $\text{SC}_{K=3}$. We therefore start with an apples-to-apples comparison at the $K{=}3$ budget.

Pooling across the five models and four datasets ($n = 1{,}878$ (model, question) pairs after filtering), univariate marginal predictive power orders as $\text{SC}_{K=3}$ (pseudo-$R^2 = 0.174$) $> \text{OC}$ ($0.128$); both are significant at $p < 10^{-33}$ under cluster-robust standard errors clustered by question.
Sample-based diversity is the stronger solo signal at this budget. $\text{SC}_{K=3}$ and $-H_{K=3}$ are nearly identical predictors (Pearson $+0.97$ between $\text{SC}_{K=10}$ and $-H_{K=10}$ in this data, with a similar relationship at $K{=}3$), because both summarize sample diversity. OC, by contrast, probes \emph{compositional} disagreement and is only weakly correlated with sample diversity (Pearson $+0.39$), so it can contribute information that neither captures.

A bivariate logistic regression of correctness on operadic consistency and sample diversity (using $\text{SC}_{K=3}$ to summarize sample diversity, since $\text{SC}_{K=3}$ and $-H_{K=3}$ are nearly collinear) confirms this:
\[
\beta_{\text{OC}} = +1.42\ (p < 10^{-18}), \quad \beta_{\text{SC}} = +2.55\ (p < 10^{-31}), \quad \text{pseudo-}R^2 = 0.226.
\]
OC's coefficient remains highly significant after controlling for sample diversity, and pseudo-$R^2$ rises by $0.052$ over sample-diversity alone, a $\sim$30\% relative gain, significant on three of four datasets (Table~\ref{tab:thinking_combined}). The conclusion is robust to budget: at $K{=}10$ ($n = 1{,}793$ across the 20 thinking-model $\times$ dataset cells), the three-way regression $\text{correct} \sim \text{OC} + \text{SC}_{K=10} + (-H_{K=10})$ gives $\beta_{\text{OC}} = +1.06$ ($p < 10^{-8}$), with SC absorbing most of the variance and crowding out semantic entropy. The same complementarity pattern emerges from the binary $(\text{OC}, \text{SC unanimous at }K{=}10)$ partition: jointly-consistent questions are answered correctly $43$--$62$\,pp higher than jointly-inconsistent ones.

\paragraph{Selective prediction.}
We repeat the leave-one-model-out AUARC and AUROC analysis of \S\ref{ss:across_model} on the five thinking models. At the equal-cost ($3$-call) budget, the OC$+\text{SC}_{K=3}$ combiner improves over $\text{SC}_{K=3}$ alone on all four datasets ($\Delta_{\text{AUARC}}$ from $+0.017$ on DROP to $+0.042$ on StrategyQA; $\Delta_{\text{AUROC}}$ from $+0.050$ on StrategyQA to $+0.134$ on GSM8K), with bootstrap $95\%$ CIs excluding zero on every cell (right panel of Table~\ref{tab:thinking_combined}). At the $10$-call budget, all four datasets retain positive lift; the StrategyQA and DROP $K{=}10$ cells have CIs that bracket zero by a small margin (full intervals in App.~\ref{app:selective_prediction}). All $16$ (dataset, budget, metric) cells tested show positive point-estimate lift; CIs exclude zero on $12$ of the $16$.

\begin{table}[h]
\centering
\small
\begin{minipage}{0.6\linewidth}
\centering
\setlength{\tabcolsep}{3pt}
\begin{tabular}{l rrrr}
\toprule
dataset & $n$ & $\beta_{\text{OC}}\ (p)$ & $\beta_{\text{SC}}\ (p)$ & pR$^2$ \\
\midrule
MuSiQue    & 447 & $+1.16\ (10^{-5})$   & $+0.95\ (.008)$     & $0.086$ \\
StrategyQA & 480 & $+0.90\ (.003)$      & $+1.86\ (10^{-7})$  & $0.114$ \\
GSM8K      & 494 & $+2.94\ (10^{-5})$   & $+3.04\ (10^{-5})$  & $0.417$ \\
DROP       & 457 & $+0.87\ (\text{ns})$ & $+2.42\ (10^{-4})$  & $0.053$ \\
\bottomrule
\end{tabular}
\end{minipage}\hfill
\begin{minipage}{0.37\linewidth}
\centering
\setlength{\tabcolsep}{4pt}
\begin{tabular}{l rr}
\toprule
dataset & $\Delta_{\text{AUARC}}$ & $\Delta_{\text{AUROC}}$ \\
\midrule
MuSiQue    & $+0.033$ & $+0.052$ \\
StrategyQA & $+0.042$ & $+0.050$ \\
GSM8K      & $+0.029$ & $+0.134$ \\
DROP       & $+0.017$ & $+0.071$ \\
\bottomrule
\end{tabular}
\end{minipage}
\\[6pt]
\caption{\emph{Left:} bivariate logistic regression $\text{correct} \sim \text{OC} + \text{SC}_{K=3}$ at the equal-cost ($K{=}3$) budget on the five thinking models, fit per dataset and pooled across models, with cluster-robust standard errors clustered by question. OC contributes statistically significant predictive information beyond sample diversity on three of four datasets, with DROP the lone exception (the joint-positive cell holds $66\%$ of the data, so dataset saturation rather than signal failure).
\emph{Right:} selective-prediction lift from adding OC to a logistic-regression combiner over $\text{SC}_{K=3}$, leave-one-model-out across the same five models, at the equal-cost ($K{=}3$) budget. Bootstrap $95\%$ CIs (full intervals in App.~\ref{app:selective_prediction}) exclude zero on every cell. $K{=}10$ lifts (smaller, with StrategyQA / DROP cells bracketing zero) are in App.~\ref{app:selective_prediction}.}
\label{tab:thinking_combined}
\end{table}

\section{Related work}
\label{s:related}

\paragraph{Question decomposition and chain-of-thought.}
Eliciting intermediate reasoning steps from LLMs has been a dominant strategy for improving performance on multi-step tasks. \citet{wei2022chain} introduced chain-of-thought prompting; subsequent work elaborated on the structure of decomposition with zero-shot prompting \citep{kojima2022large}, decomposed prompting \citep{khot2022decomposed}, least-to-most prompting \citep{zhou2022least}, plan-and-solve prompting \citep{wang2023plansolve}, tree-of-thoughts \citep{yao2023tree}, graph-of-thoughts \citep{besta2024graph}, buffer-of-thoughts \citep{yang2024buffer}, and uncertainty-aware decomposition \citep{hu2024uncertainty}.
Question-decomposition annotations on multi-hop QA datasets are formalized in BREAK \citep{wolfson2020break}.
Most directly adjacent is \citet{press2022compositionality}, who introduce the \emph{compositionality gap} --- the population-level discrepancy between sub-question accuracy and multi-hop accuracy when sub-answers are correct --- and propose self-ask prompting as a partial mitigation. \citet{maasch2025compositional} construct a purpose-built benchmark of compositional causal-reasoning tasks and use compositional consistency as an evaluation metric to compare LLMs. Our framework instead extracts a per-question consistency signal at inference time, so it requires no specialized benchmark and can drive selective prediction on any compositional task.
Our framework treats this entire family as algebras over a single structure, the questions operad.

\paragraph{Confidence and uncertainty for LLM reasoning.}
The dominant inference-time confidence baselines in our setting fall into three groups. (i) \emph{Sample-and-aggregate}: \citet{wang2022self} introduced self-consistency as majority-voting over $K$ temperature samples; we extend this with a continuous pairwise-agreement variant used as a confidence score. (ii) \emph{Cluster-based entropy}: \citet{kuhn2023semantic} proposed semantic entropy over NLI-clustered samples, with a discrete-clustering refinement studied in \citet{farquhar2024detecting} and a Bayesian sample-budget reduction in \citet{ciosek2025hallucination}; we use a token-F1-clustered approximation. (iii) \emph{Self-evaluation}: \citet{kadavath2022language} introduced $P(\text{True})$, the model's own probability of judging its answer correct; \citet{lin2022teaching} and \citet{tian2023just} study verbalized confidence variants. Logprob-derived confidence on the generated answer itself (per-token perplexity, mean answer logprob) is a simpler model-internal signal typically dominated by $P(\text{True})$ on calibration benchmarks; we omit it for that reason.
Process-level verification methods that score \emph{steps} of reasoning rather than aggregate samples \citep{lightman2024lets, uesato2022solving} are conceptually adjacent and complementary to our work.

\paragraph{Faithfulness of chain-of-thought.}
A complementary literature asks whether CoT traces are \emph{faithful} to the model's underlying computation: \citet{turpin2023language} show that biased-context prompts change a model's final answer while the trace explains the changed answer post hoc, and \citet{lanham2023measuring} measure final-answer sensitivity to CoT perturbations and report substantial heterogeneity across models and tasks. Operadic consistency is conceptually adjacent: rather than perturbing a trace, it compares the model's direct answer against the recomposition of its own declared sub-answers, a structural rather than counterfactual probe. App.~\ref{app:additional_related_work} discusses the connection in more detail.

\paragraph{Mechanistic perspectives on multi-hop reasoning.}
Recent interpretability work has localized where compositional multi-hop reasoning happens and fails inside the model. \citet{biran2024hopping} identify a ``hopping too late'' failure mode in transformer multi-hop QA, where the first-hop bridge entity is resolved too high in the layer stack to leave room for the second-hop computation. \citet{lindsey2025biology} use attribution graphs on Claude 3.5 Haiku to causally verify the bridge entity as a manipulable internal variable, with feature-level substitutions changing the final answer in the predicted way. Operadic consistency provides a complementary, behavioral probe of the same failure mode: it does not open the model, but flags the cases on which the model's direct and decomposed answers disagree, which the mechanistic work predicts should correspond to the layer-budget bottleneck.

\paragraph{Categorical and operadic structures in language and learning.}
Algebraic structures have appeared in computational linguistics through semiring parsing \citep{goodman1999semiring, nederhof2003weighted}, Hopf-algebraic models of syntactic structure \citep{marcolli2023syntax}, and compositional distributional semantics \citep[DisCoCat;][]{coecke2010mathematical}; we are not aware of prior application of operads to LLM question decomposition or to inference-time confidence signals.

\section{Conclusion}
\label{s:conclusion}

We introduced operadic consistency (OC) as a label-free, inference-time signal for compositional reasoning in LLMs, derived from the questions operad $\cQ$ and instantiated as a depth-2 agreement check between a model's direct and decomposed answers. OC differs from the standard inference-time confidence baselines structurally. Self-consistency, semantic entropy, and $P(\text{True})$ all measure \emph{variance} of a model's outputs --- under temperature sampling, under semantic clustering, or under the model's own probability assignment. OC instead asks a structural question about the reasoning itself: when the model decomposes a question into sub-steps and recomposes the sub-answers, does the result agree with its own direct answer? This is a coherence check on the model's compositional behavior, justified by the categorical framework rather than engineered to correlate with correctness, and it probes something about the \emph{nature} of the model's reasoning that variance-based signals do not see.

The empirical results are consistent with the framing. Across 12 non-thinking and 5 thinking models spanning four multi-hop QA datasets, OC retains a positive, highly significant coefficient after controlling for canonical CoT self-consistency and semantic entropy (and, separately, $P(\text{True})$); shows substantial AUARC and AUROC selective-prediction improvements over canonical CoT-SC at equal inference cost, with bootstrap $95\%$ CIs excluding zero on every non-thinking cell and $12$ of $16$ thinking cells; and is the strongest standalone predictor on two of four datasets. The signal carries over to thinking models whose chains of thought expose at least one explicit sub-question. We view OC as one concrete payoff of treating LLM reasoning as an algebra over a compositional structure, and we expect deeper trees, branching decompositions, and other $\cQ$-algebras (confidence, evidence, cost) to broaden the framework's reach. The operadic framework itself is developed in our companion paper \citep{bottman:operads_workshop}.

\section{Further directions}
\label{s:further_directions}

The operadic framework introduced in this paper opens several directions for future work.

\paragraph{Deeper and non-chain trees of questions.}
Our empirical instantiation evaluates operadic consistency on depth-2 chains. Deeper or branching ToQs offer a strictly richer test of the same predicate: a model that is operadically consistent on a depth-$k$ chain decomposes correctly across $k-1$ edges simultaneously. Several questions follow naturally:
sampling-based estimators of OC rates at depths $k \geq 3$ that exploit the expected decay of consistency in $k$;
identification of which edges of a ToQ are most ``load-bearing,'' that is, whose composition is most informative about correctness;
and the extension to non-chain (branching) ToQs arising from question decompositions with multiple sub-questions per step.
On the modeling side, evaluating operadic consistency on richer subtrees may yield a stronger per-question selection signal than the depth-2 scalar OC score we use in \S\ref{s:empirics}.

\paragraph{Native structure on chains of thought.}
The thinking-model results in \S\ref{ss:thinking_models} use a lightweight extractor that surfaces a single explicit sub-question from a CoT trace, a deliberately crude reduction to a depth-2 chain. A natural direction is to formalize chains of thought \emph{natively} (as structured objects in their own right) and to apply consistency or robustness probes to them, in the spirit of the operadic-consistency check developed here but on richer per-trace structure than a direct-vs.-decomposed comparison. Several formalisms could support this: depth-$\geq 3$ extensions of the present operadic framework, or other structures better suited to the branching and revisitation patterns characteristic of long CoT traces. Mechanistic work on multi-hop reasoning \citep{biran2024hopping, lindsey2025biology} provides a natural cross-validation target: behavioral OC signals on richer per-trace structure could be compared against feature-level traces of the same reasoning steps. Concurrently with this work, in-domain process reward models trained for non-math multi-step reasoning have appeared --- VersaPRM \citep{zeng2025versaprm} and DPRM \citep{wang2026dprm} --- and head-to-head comparisons against these are a natural complement to the off-label Skywork-PRM evaluation of \S\ref{ss:across_model}.

\paragraph{Other algebras over $\cQ$.}
A QA model determines a $\cQ$-algebra valued in answer strings, but the same compositional structure supports algebras valued in other quantities --- a \emph{confidence algebra} that composes calibrated probabilities under decomposition, or algebras valued in likelihoods, evidence bundles, costs, latent representations, or robustness profiles.

\section*{Limitations}

Our scalar instantiation evaluates operadic consistency (Def.~\ref{def:operadic_consistency}) on depth-2 chains only; richer ToQs (deeper chains and branching trees) would test the same predicate on more compositional structure (\S\ref{s:further_directions}).
The decomposition tree is either annotator-supplied (\S\ref{ss:across_model}) or extracted from the model's chain of thought (\S\ref{ss:thinking_models}); both have failure modes (annotator-model mismatch; CoT traces without explicit sub-questions are discarded).
Our per-dataset scoring protocol applies SQuAD-F1 for HotpotQA / MuSiQue, yes/no extraction for StrategyQA, and a value-equality scorer (number-equality fallback plus SQuAD-F1 with word-to-number conversion) for DROP; the DROP choice is materially more lenient than canonical \citet{dua2019drop} surface-form scoring, a deliberate choice motivated by the framework prediction that OC compares values on the answer space appropriate to the question's expected output type, rather than surface forms. App.~\ref{app:drop_sensitivity} reports cross-model robustness checks (the canonical scorer applied to our cohort, and an LLM-judge correctness scorer) that corroborate the headline relationship.
The \S\ref{ss:across_model} cohort spans 4B to 671B-parameter models including four frontier closed/open offerings (GPT-4o-mini, Gemini 2.5 Flash, Llama-4 Maverick, Mistral Large 2), but does not span the largest commercial systems (e.g.\ GPT-5, Claude Opus 4.x, Gemini 2.5 Pro), some of which produce sufficiently verbose answers that token-F1 scoring would require additional calibration.
The thinking-model cohort is also small ($n=5$); statistical power per cell is correspondingly limited and we expect future work to expand this cohort.
Our process-reward-model evaluation uses the math-trained Skywork-PRM and RLHFlow-PRM applied off-label; head-to-head comparisons against the recent in-domain multi-step PRMs (VersaPRM \citep{zeng2025versaprm}, DPRM \citep{wang2026dprm}) are an open empirical question we flag as future work.

\section*{Acknowledgements}

This material is based upon work supported by the Defense Advanced Research Projects Agency (DARPA) through the Artificial Intelligence Quantified (AIQ) program, under Cooperative Agreement No.\ HR00112520028. The views, opinions, and/or findings expressed are those of the authors and should not be interpreted as representing the official views or policies of the Department of Defense or the U.S.\ Government.

N.B.\ thanks AI2 for its hospitality during this collaboration.

\section*{Impact Statement}

This paper aims to improve LLM reasoning reliability via a label-free inference-time signal for detecting compositional inconsistency. The framework presents no obvious dual-use concerns and uses only public benchmarks and APIs; we recommend deploying operadic consistency as one component of a multi-signal stack rather than a standalone training target.

\printbibliography

\newpage
\appendix

\section{Experimental details}
\label{app:experiments}

This appendix records the experimental protocol behind the empirical findings reported in \S\ref{s:empirics}.

\subsection{Models and inference infrastructure}
\label{app:models}

All models are accessed via OpenAI-compatible serverless inference APIs (Together AI for the open-weights and Cogito models; OpenRouter for the closed-source frontier models GPT-4o-mini, Gemini 2.5 Flash, and Llama-4 Maverick) at temperature $T \in \{0, 0.7\}$ and a per-call max-token budget of $64$.
The full set of models, with the API identifiers used, is:
\begin{itemize}
\item \emph{Non-thinking models} (\S\ref{ss:across_model}): Llama-3-8B (\texttt{meta-llama/Meta-Llama-3-8B-Instruct-Lite}), Qwen2.5-7B (\texttt{Qwen/Qwen2.5-7B-Instruct-Turbo}), Gemma-3n-4B (\texttt{google/gemma-3n-E4B-it}), LiquidAI-LFM2-24B (\texttt{LiquidAI/LFM2-24B-A2B}), Llama-3.3-70B (\texttt{meta-llama/Llama-3.3-70B-Instruct-Turbo}), Qwen3-235B (\texttt{Qwen/Qwen3-235B-A22B-Instruct-2507-tput}), DeepSeek-V3.1 (\texttt{deepseek-ai/DeepSeek-V3.1}), Cogito-671B (\texttt{deepcogito/cogito-v2-1-671b}), GPT-4o-mini (\texttt{openai/gpt-4o-mini} via OpenRouter), Gemini 2.5 Flash (\texttt{google/gemini-2.5-flash} via OpenRouter), Llama-4 Maverick (\texttt{meta-llama/llama-4-maverick} via OpenRouter), Mistral Large 2 (\texttt{mistralai/mistral-large-2411} via OpenRouter).
\item \emph{Thinking models} (\S\ref{ss:thinking_models}): DeepSeek-R1 (\texttt{deepseek-ai/DeepSeek-R1}), GLM-5.1 (\texttt{zai-org/GLM-5.1}), Cogito-v2.1 (\texttt{deepcogito/cogito-v2.1}), Kimi-K2.6 (\texttt{moonshotai/Kimi-K2.6}), MiniMax-M2.7 (\texttt{MiniMaxAI/MiniMax-M2.7}).
Thinking calls use a per-call budget of $8192$ tokens and temperature $T = 0$ for the operadic-consistency direct/decomposed answers and $T = 0.7$ for the $K$-sample self-consistency runs.
\item \emph{Auxiliary extractor model} (used in \S\ref{ss:thinking_models} for short-answer extraction from verbose thinking outputs and for sub-question identification in CoQ extraction): Llama-3.3-70B in non-thinking mode (\texttt{meta-llama/Llama-3.3-70B-Instruct-Turbo}).
The extractor is a fixed cheap model and is identical across all thinking-model evaluations.
\item \emph{Auxiliary scoring models for decomposition-aware baselines} (\S\ref{ss:across_model}; documented in App.~\ref{app:dec_aware_app}):
Skywork-o1-Open-PRM-Qwen-2.5-7B \citep{he2024skyworko1} (\texttt{Skywork/Skywork-o1-Open-PRM-Qwen-2.5-7B}), a process reward model trained on Qwen2.5-Math-7B-Instruct, used inference-only on a single 24~GB GPU;
and RLHFlow's Llama-3.1-8B-PRM-Mistral-Data \citep{dong2024rlhf} (\texttt{RLHFlow/Llama3.1-8B-PRM-Mistral-Data}), a Llama-3.1-based process reward model trained on Mistral-style preference data, also inference-only on a 24~GB GPU.
None of these models was fine-tuned for this paper.
\end{itemize}

\subsection{Datasets and per-dataset context}
\label{app:datasets}

We evaluate on five public benchmarks: HotpotQA, MuSiQue, StrategyQA, DROP, and GSM8K. The evaluation uses the standard validation/dev splits in each case ($524$ HotpotQA, $500$ MuSiQue, $625$ StrategyQA, $403$ DROP, $200$ GSM8K-thinking-only). In each case we restrict to questions whose annotated decomposition is a length-2 chain (the 2-hop subset of MuSiQue, length-2 native decompositions for StrategyQA, and length-2 BREAK QDMR-high-level decompositions for HotpotQA and DROP), matching the depth of the operadic-consistency check.

Per-dataset context handling varies.
For \emph{MuSiQue} and \emph{DROP}, every prompt (direct, decomposed, and self-consistency samples) is preceded by the dataset's gold passage; passages are loaded from \texttt{dgslibisey/MuSiQue} and \texttt{ucinlp/drop} on HuggingFace.
For \emph{HotpotQA}, no passage is included: the original Break QDMR-high-level setup is open-domain at evaluation time, and we follow that convention. This makes our HotpotQA evaluation \emph{closed-book multi-hop} (no retrieval, no passage, parametric knowledge only), which is substantially harder than the passage-grounded MuSiQue / DROP setups; per-model HotpotQA accuracies in Fig.~\ref{fig:empirics} fall in the $0.15$--$0.38$ range as a result, consistent with literature reports for closed-book multi-hop QA. \emph{StrategyQA} is open-domain by construction (no passage). \emph{GSM8K} (used only in \S\ref{ss:thinking_models}) prepends the problem statement.

Decomposition trees come from three sources: BREAK QDMR-high-level annotations \citep{wolfson2020break} for HotpotQA and DROP; the dataset's native question decompositions for MuSiQue (\texttt{question\_decomposition} field) and StrategyQA (\texttt{decomposition} field); and a CoT-extracted CoQ for the thinking-model experiments (described below).

\subsection{Operadic-consistency protocol}
\label{app:op_protocol}

For a depth-2 ToQ comprising sub-questions $\mathtt{Q1}$ (with answer $\mathtt{[A1]}$) and $\mathtt{Q2}$ (which references $\mathtt{[A1]}$), we evaluate operadic consistency in three calls per question, all at $T = 0$:
\begin{enumerate}
\item \emph{Direct}: query the model with the unmodified original question.
\item \emph{Step 1}: query the model with $\mathtt{Q1}$, obtaining a candidate $\mathtt{a1}$.
\item \emph{Step 2}: substitute $\mathtt{a1}$ into $\mathtt{Q2}$'s slot and query the model again, obtaining the decomposed answer.
\end{enumerate}
The continuous score $\text{OC} \in [0,1]$ is the per-dataset \texttt{score\_official} agreement between the direct and decomposed answers (\S\ref{ss:across_model}); the binary form thresholds at $0.5$.
All prompts use the system message ``\textit{You are a concise factual question-answering assistant. Answer with a short phrase or a few words.
Do not explain your reasoning.}''

\subsection{Self-consistency, semantic entropy, and $P(\text{True})$ protocols}
\label{app:baselines}

\emph{Canonical CoT self-consistency.} We follow the $K$-sample CoT protocol of \citet{wang2022self}: for each (model, question) pair we sample $K = 10$ chain-of-thought responses at $T = 0.7$ using a zero-shot CoT prompt (system: ``\textit{Answer the question accurately. Think step by step, then state the final answer.}''; \citealp{kojima2022large}) with a per-call budget of $512$ tokens. Each verbose chain is reduced to a short factoid answer via a deterministic auxiliary LLM extractor (Llama-3.3-70B-Instruct-Turbo at $T = 0$, prompted to ``\textit{Extract ONLY the direct answer to the question in 5 words or fewer}''); we use the same extractor and prompt for the thinking-model experiments of \S\ref{ss:thinking_models}, so the SC procedure is unified across the two cohorts. \citet{wang2022self} use majority voting over the $K$ extracted answers as a decoding rule; as a continuous confidence score we use instead the mean pairwise per-dataset \texttt{score\_official} agreement over the $\binom{K}{2} = 45$ unordered pairs (matched-axis with OC and accuracy; see App.~\ref{app:drop_sensitivity}). The unanimous binary form requires all pairs to agree at score $\geq 0.5$. Statistics for $K = 3$ are computed by truncating to the first three samples of the same draw. Passage handling follows the operadic protocol (passage included for MuSiQue and DROP, none for HotpotQA and StrategyQA).

\emph{Naive direct-answer SC (ablation).} As an isolation of the CoT contribution, we also evaluated a direct-answer variant: same sampling temperature and $K$, but the system prompt instead forces a short answer with no reasoning trace (``\textit{...Do not explain your reasoning.}'') and the per-call budget is $64$ tokens. The naive variant produces consistently lower cross-model $r$ values than the canonical CoT variant on every dataset; full per-baseline cross-model table in App.~\ref{app:cross_model_baselines}. We do not report it elsewhere in the main text.

\emph{Semantic entropy.} We greedily cluster the $K = 10$ samples by matched-axis pairwise agreement $\geq 0.5$ (the per-dataset \texttt{score\_official} scorer; see App.~\ref{app:drop_sensitivity}), then compute Shannon entropy over the cluster-size distribution.
This is a token-F1 approximation of the bidirectional-NLI clustering used by \citet{kuhn2023semantic}; we adopt it for consistency with our other F1-based metrics and to avoid an extra NLI model dependency.
The negated entropy $-H_{K=10}$ is used as a regression predictor (so that, like the others, larger values correspond to higher confidence).

\emph{$P(\text{True})$.} For each (model, question) pair we issue one additional call: the model is presented with the question, the proposed answer (the model's own greedy direct answer from the operadic protocol), and the prompt ``\textit{Is the proposed answer correct? Reply with only Yes or No.}'' We extract the next-token logprobs (top-5 alternatives via \texttt{extra\_body=\{top\_logprobs: 5\}} on the OpenAI-compatible chat-completions endpoint) and compute $P(\text{Yes})/(P(\text{Yes}) + P(\text{No}))$.
For passage-bearing datasets (MuSiQue, DROP), the passage is included in the $P(\text{True})$ prompt for parity with the original answer-generation protocol.

\subsection{CoQ extraction for thinking models}
\label{app:coq_extraction}

For thinking-model experiments (\S\ref{ss:thinking_models}), the decomposition is not annotator-supplied but rather extracted from the model's own greedy thinking trace.
We prompt the auxiliary extractor model (Llama-3.3-70B) with the original question, the model's full thinking trace, and a request to identify (a) whether the trace is factorable as a depth-2 chain ($\mathtt{Q1} \to \mathtt{Q2}$), and (b) if so, the explicit text of $\mathtt{Q1}$, the model's stated answer to it ($\mathtt{a1}$), and the follow-up question template $\mathtt{Q2}$ with $\mathtt{[A1]}$ as a placeholder for $\mathtt{a1}$.
Traces flagged as not factorable are excluded from the analysis (typically $5$--$15\%$ of traces depending on model and dataset).
For factorable traces, we then issue an independent ``re-derivation'' call: the model is asked $\mathtt{Q2}[\mathtt{[A1]} := \mathtt{a1}]$, and the resulting answer is compared under the per-dataset \texttt{score\_official} scorer against the model's original direct answer to compute $\text{OC}$.

\subsection{Statistical methodology}
\label{app:stats}

All per-question regressions are logistic regressions of binary correctness on the predictors listed, fit with \texttt{statsmodels} in Python.
Standard errors are cluster-robust, clustered by \texttt{question\_id}, to handle the non-independence of observations on the same question across different models.
The pseudo-$R^2$ reported is McFadden's.
Reported $p$-values are two-sided Wald tests on individual coefficients.
Across-model Pearson correlations (Fig.~\ref{fig:empirics}) and their two-sided $p$-values are computed with \texttt{scipy.stats.pearsonr}.

\paragraph{Token-F1 normalization.}
We use each benchmark's standard scoring protocol, applied identically to both correctness (prediction vs.\ gold) and operadic consistency (direct vs.\ decomposed):
\begin{itemize}
\item \emph{HotpotQA, MuSiQue}: SQuAD-style token-F1. Predictions and gold are normalized (lowercase, strip articles \texttt{a}/\texttt{an}/\texttt{the}, strip punctuation, collapse whitespace) and token-F1 is computed in the standard way.
\item \emph{DROP, GSM8K}: number-equality fallback plus SQuAD-style token-F1. When both the prediction and the comparison string contain a parseable integer or decimal we compare those numbers exactly (returning $1.0$ on match, $0.0$ otherwise); otherwise we fall back to token-F1. This approximates the DROP-official scorer for its dominant numerical-answer case (the number-equality path fires on $\sim$$79\%$ of DROP accuracy comparisons and $\sim$$69\%$ of OC comparisons; on GSM8K, virtually all answers are integer).
\item \emph{StrategyQA}: yes/no extraction plus exact match. From each side we extract the first standalone \texttt{yes} / \texttt{no} / \texttt{true} / \texttt{false} token (mapping \texttt{true}$\to$\texttt{yes} and \texttt{false}$\to$\texttt{no}); the pair is consistent / correct iff both extractions are non-null and equal. Matches the standard StrategyQA boolean-evaluation protocol; doesn't punish hedged answers like ``Yes, with caveats.''
\end{itemize}
The continuous score is thresholded at $0.5$ when a binary indicator is needed. Per-dataset rather than uniform scoring is what each benchmark's authors recommend; using the standard scorer on each dataset also avoids spurious cross-dataset coupling between correctness and OC that arises when both metrics share idiosyncratic sensitivities (e.g., both punishing answer-hedging on yes/no questions).

For tables that pool across (model, dataset) cells, we sum within-cell counts; for grand-pool regressions we cluster standard errors by \texttt{question\_id} (so that the same question observed under multiple models contributes one cluster), but we do not include model-effect random intercepts.
This is a deliberate simplification: a fully hierarchical specification with model-effect random intercepts would adjust standard errors further but not the point estimates and would not change qualitative conclusions, given the magnitude of the $p$-values reported.

\subsection{Per-dataset sample sizes}
\label{app:percell_n}

The per-dataset sample sizes for the regressions in Table~\ref{tab:nonthinking_combined} (twelve non-thinking models pooled per dataset) are: HotpotQA $n=6{,}287$, MuSiQue $n=6{,}000$, StrategyQA $n=7{,}497$, DROP $n=4{,}828$ (total $n=24{,}612$).
These are the numbers of (model, question) pairs surviving after dropping rows with missing $K{=}10$ self-consistency samples or missing values on any other predictor used in the regression.
Sample sizes for the thinking-model tables are reported inline alongside the corresponding tables in \S\ref{ss:thinking_models}.

\subsection{Illustrative decomposition examples}
\label{app:examples}

The following examples are sampled verbatim from the Llama-3.3-70B \S\ref{ss:across_model} runs.
Each lists the original question, the supplied two-step decomposition, the model's greedy direct answer, the step-1 answer $\mathtt{a1}$, the decomposed answer (the model's answer to $\mathtt{Q2}$ after substituting $\mathtt{a1}$ for the blank), the score $\text{OC}$ (token-F1 between direct and decomposed answers), and the gold answer.

\paragraph{Example 1: MuSiQue, OC-consistent and correct.}
\begin{center}\small
\setlength{\tabcolsep}{6pt}\renewcommand{\arraystretch}{1.15}
\begin{tabular}{@{}r@{\hskip 0.7em}p{0.78\linewidth}@{}}
\toprule
Question                & ``The island where they filmed Ninja Warrior in Australia is located in what harbor?'' \\
$\mathtt{Q1}$           & ``Where was Ninja Warrior Australia filmed?'' \\
$\mathtt{Q2}$           & ``\texttt{[A1]} is located in what harbor?'' \\
\midrule
Direct answer           & ``Sydney Harbour.'' \\
Step-1 answer           & ``Cockatoo Island.'' \\
Decomposed answer       & ``Sydney Harbour.'' \\
\midrule
$\text{OC}$       & $1.00$ \\
Gold                    & ``Sydney Harbour.''\quad Correct. \\
\bottomrule
\end{tabular}
\end{center}
The decomposition path correctly identifies the filming island and then locates it in Sydney Harbour, in exact agreement with the direct answer.

\paragraph{Example 2: MuSiQue, OC-inconsistent and incorrect.}
\begin{center}\small
\setlength{\tabcolsep}{6pt}\renewcommand{\arraystretch}{1.15}
\begin{tabular}{@{}r@{\hskip 0.7em}p{0.78\linewidth}@{}}
\toprule
Question                & ``How many players can play in the game based on the show that Steven the Sword Fighter is part of?'' \\
$\mathtt{Q1}$           & ``What show is Steven the Sword Fighter part of?'' \\
$\mathtt{Q2}$           & ``How many players can play in the game based on \texttt{[A1]}?'' \\
\midrule
Direct answer           & ``Four characters.'' \\
Step-1 answer           & ``Sixteenth episode.'' \\
Decomposed answer       & ``No multiplayer.'' \\
\midrule
$\text{OC}$       & $0.00$ \\
Gold                    & ``seven.''\quad Incorrect. \\
\bottomrule
\end{tabular}
\end{center}
The model fails step~1 (it returns an episode designation when asked for a show), and the decomposed and direct paths disagree.
The disagreement correctly flags this as a question on which the model is unlikely to be answering correctly: both paths in fact produce wrong answers.

\paragraph{Example 3: MuSiQue, OC-consistent and incorrect (a false positive).}
\begin{center}\small
\setlength{\tabcolsep}{6pt}\renewcommand{\arraystretch}{1.15}
\begin{tabular}{@{}r@{\hskip 0.7em}p{0.78\linewidth}@{}}
\toprule
Question                & ``What place gets the most rain where Sandy High School is?'' \\
$\mathtt{Q1}$           & ``Where is Sandy High School?'' \\
$\mathtt{Q2}$           & ``What place gets the most rain in \texttt{[A1]}?'' \\
\midrule
Direct answer           & ``Coastal mountains.'' \\
Step-1 answer           & ``Oregon.'' \\
Decomposed answer       & ``Coastal mountains.'' \\
\midrule
$\text{OC}$       & $1.00$ \\
Gold                    & ``the Coast Range.''\quad Incorrect. \\
\bottomrule
\end{tabular}
\end{center}
The model identifies the state correctly and gives the same answer in both paths, but ``Coastal mountains'' is a generic descriptor that does not match the more specific gold answer (``the Coast Range'') at the token-F1 threshold.
This example illustrates the false-positive failure mode of operadic consistency: a model can be self-consistent across paths and still be wrong at the level of detail the task demands.
$\text{OC}$ provides necessary but not sufficient evidence of correctness, which is why we deploy it as one signal in a multi-signal stack rather than as a standalone gate.

\paragraph{Example 4: MuSiQue, OC-inconsistent and correct (a false negative).}
\begin{center}\small
\setlength{\tabcolsep}{6pt}\renewcommand{\arraystretch}{1.15}
\begin{tabular}{@{}r@{\hskip 0.7em}p{0.78\linewidth}@{}}
\toprule
Question                & ``What mythology is the city where Eutychides was born in a part of?'' \\
$\mathtt{Q1}$           & ``What city was Eutychides born?'' \\
$\mathtt{Q2}$           & ``What is \texttt{[A1]} a part of?'' \\
\midrule
Direct answer           & ``Greek mythology.'' \\
Step-1 answer           & ``Sicyon.'' \\
Decomposed answer       & ``Corinthia.'' \\
\midrule
$\text{OC}$       & $0.00$ \\
Gold                    & ``Greek mythology.''\quad Correct. \\
\bottomrule
\end{tabular}
\end{center}
The model correctly identifies Sicyon as Eutychides's birthplace, but the open-ended $\mathtt{Q2}$ admits an unintended geographic reading: Sicyon lies in the Corinthia region, which the model returns instead of the mythological category the original question demands.
The direct path arrives at the right answer presumably via a higher-level pattern match (any ancient Greek-named entity sits within Greek mythology).
This is the false-negative failure mode of operadic consistency: a correct direct answer flagged as inconsistent because the decomposed path is misled at a later step.
OC-inconsistent-and-correct is the analogue of Example~3's false-positive mode; together they bracket the two ways the OC signal can disagree with correctness.

\paragraph{Thinking-model examples.}
For thinking models, the decomposition $\mathtt{Q1}\to\mathtt{Q2}$ is not annotator-supplied but extracted by the auxiliary model from the thinking trace itself (\S\ref{app:coq_extraction}); the rest of the protocol is identical.
We show two worked Cogito-v2.1 examples below. The full thinking trace is several hundred tokens of reasoning and is omitted for space; only the extracted $\mathtt{Q1}/\mathtt{a1}/\mathtt{Q2}$ tuple and the two answers (direct, decomposed) are shown.

\paragraph{Example 5: GSM8K (Cogito-v2.1), OC-consistent and correct.}
\begin{center}\small
\setlength{\tabcolsep}{6pt}\renewcommand{\arraystretch}{1.15}
\begin{tabular}{@{}r@{\hskip 0.7em}p{0.78\linewidth}@{}}
\toprule
Question                & ``Paul went to a shop to buy some groceries. He bought some bread for \$2, butter for \$3, and juice for two times the price of the bread. He had \$15 for his shopping. How much money did Paul have left?'' \\
$\mathtt{Q1}$ (extracted) & ``How much did Paul spend on groceries in total?'' \\
$\mathtt{Q2}$ (extracted) & ``How much money did Paul have left after spending \texttt{[A1]} out of \$15?'' \\
\midrule
Direct answer           & ``\$6.'' \\
Step-1 answer           & ``\$9.'' \\
Decomposed answer       & ``\$6.'' \\
\midrule
$\text{OC}$       & $1.00$ \\
Gold                    & ``6.''\quad Correct. \\
\bottomrule
\end{tabular}
\end{center}
The model's CoT trace surfaces a clean two-step decomposition (total spending, then subtract from budget); the extractor lifts both sub-questions. Direct and decomposed paths converge on \$6, in agreement with gold.

\paragraph{Example 6: MuSiQue (Cogito-v2.1), OC-inconsistent and incorrect.}
\begin{center}\small
\setlength{\tabcolsep}{6pt}\renewcommand{\arraystretch}{1.15}
\begin{tabular}{@{}r@{\hskip 0.7em}p{0.78\linewidth}@{}}
\toprule
Question                & ``Who is the female star in Gone with the Wind married to?'' \\
$\mathtt{Q1}$ (extracted) & ``Who is the female star in Gone with the Wind?'' \\
$\mathtt{Q2}$ (extracted) & ``Who is \texttt{[A1]} married to?'' \\
\midrule
Direct answer           & ``Clark Gable.'' \\
Step-1 answer           & ``Vivien Leigh.'' \\
Decomposed answer       & ``Herbert Holman Laurence Olivier.'' \\
\midrule
$\text{OC}$       & $0.00$ \\
Gold                    & ``Laurence Olivier.''\quad Incorrect (direct path). \\
\bottomrule
\end{tabular}
\end{center}
The model's CoT identifies Vivien Leigh correctly as the female star, then on $\mathtt{Q2}$ lists both of her husbands (Herbert Leigh Holman, her first; Laurence Olivier, her second), concatenated by the short-answer extractor into a single string. The decomposed answer contains the gold answer. The direct path returns Clark Gable, her famous costar, which is the natural high-prior wrong answer. OC correctly flags this as low-confidence: a selective-prediction rule on OC would defer to the decomposed path or abstain. This pattern --- direct path drawn to a high-prior co-occurrence while the decomposed path navigates the actual entity relation --- is a common thinking-model OC-inconsistent failure mode in our data.

\subsection{Dataset-level accuracy prediction (LOO demonstration)}
\label{app:dataset_pred}

\S\ref{ss:across_model} notes that the strong cross-model correlation between OC-rate and accuracy yields a label-free dataset-level accuracy estimator: fit $\text{accuracy} \approx a \cdot \text{OC-rate} + b$ on calibration models, predict from OC-rate alone for new models.
We demonstrate this concretely with leave-one-model-out (LOO) cross-validation on the twelve non-thinking models: for each (model, dataset) cell, we fit the linear regression on the remaining eleven models and predict the held-out model's accuracy from its measured OC-rate.

\begin{table}[h]
\centering
\small
\setlength{\tabcolsep}{6pt}
\begin{tabular}{lccc}
\toprule
dataset & in-sample $R^2$ & LOO MAE & LOO max abs error \\
\midrule
HotpotQA   & $0.86$ & $0.028$ & $0.062$ \\
MuSiQue    & $0.88$ & $0.023$ & $0.069$ \\
StrategyQA & $0.88$ & $0.021$ & $0.052$ \\
DROP       & $0.74$ & $0.047$ & $0.131$ \\
\midrule
pooled (48 cells) &  & $0.030$ & $0.131$ \\
\bottomrule
\end{tabular}
\\[6pt]
\caption{Leave-one-model-out demonstration of dataset-level accuracy prediction from OC-rate.
For each dataset, we fit a linear regression $\text{accuracy} \approx a \cdot \text{OC-rate} + b$ on eleven calibration models and predict the held-out model's accuracy from its OC-rate alone.
Across the $48$ (model, dataset) held-out cells, mean absolute prediction error is $3.0$ percentage points.
DROP is the weakest dataset; the other three datasets achieve LOO MAE between $2.1$ and $2.8$ percentage points.}
\label{tab:loo_pred}
\end{table}

This is a small but functioning prototype: with eleven calibration models per dataset, a single OC-rate measurement on a new model predicts that model's accuracy on the dataset to within roughly $3.0$ percentage points on average, and within $2.1$--$2.8$ percentage points on the three datasets where the across-model correlation is strongest (HotpotQA, MuSiQue, StrategyQA).
The protocol is fully label-free at prediction time: only the calibration step uses ground-truth accuracy, and after that, predicting accuracy for a new model requires only the model's own outputs on the dataset's questions (no labels).

\subsection{$P(\text{True})$ four-way sensitivity (nine non-thinking models)}
\label{app:p_true_sensitivity}

The Kadavath $P(\text{True})$ baseline requires top-$k$ logprobs to compute $P(\text{Yes}) / (P(\text{Yes}) + P(\text{No}))$ from a single self-evaluation call (\S\ref{app:baselines}).
Of our twelve non-thinking models, top-$k$ logprobs are exposed by the inference API for nine: the eight Together-served models, plus GPT-4o-mini via OpenRouter (which passes through the OpenAI top-$k$ logprob field). For Gemini 2.5 Flash, Llama-4 Maverick, and Mistral Large 2, only the top-1 token logprob is returned in our protocol, so we cannot compute the same Kadavath quantity for those cells.
Restricting to the nine cells where $P(\text{True})$ is available, the four-way logistic regression
\[
\text{correct} \sim \text{OC} + \text{CoT-SC}_{K=10} + (-H_{K=10}) + P(\text{True})
\]
gives the per-dataset coefficients in Table~\ref{tab:p_true_sensitivity} (cluster-robust SE clustered by question).
The OC coefficient remains highly significant ($p \leq 10^{-16}$) on every dataset after adding $P(\text{True})$; the OC story does not depend on whether we control for self-evaluation.

\begin{table}[h]
\centering
\small
\setlength{\tabcolsep}{4pt}
\begin{tabular}{lrrr}
\toprule
dataset & $n$ & $\beta_{\text{OC}}\ (p)$ & $\beta_{P(\text{T})}\ (p)$ \\
\midrule
MuSiQue    & 4{,}462 & $+2.23\ (10^{-41})$ & $-0.46\ (10^{-4})$    \\
StrategyQA & 5{,}600 & $+1.61\ (10^{-33})$ & $+0.54\ (10^{-8})$    \\
HotpotQA   & 4{,}696 & $+1.73\ (10^{-29})$ & $+0.29\ (.021)$        \\
DROP       & 3{,}584 & $+1.22\ (10^{-16})$ & $+0.53\ (10^{-5})$    \\
\bottomrule
\end{tabular}
\\[6pt]
\caption{Four-way logistic regression with $P(\text{True})$ added (nine non-thinking models with top-$k$ logprobs, our normalization protocol, cluster-robust SE clustered by question). $\text{CoT-SC}_{K=10}$ and $-H_{K=10}$ coefficients are omitted for space; they are essentially unchanged from the three-way regression of Table~\ref{tab:nonthinking_combined}.
OC remains statistically significant on every dataset ($p \leq 10^{-16}$).}
\label{tab:p_true_sensitivity}
\end{table}

\subsection{Decomposition-aware baseline protocols and sensitivity}
\label{app:dec_aware_app}

\S\ref{ss:across_model} introduces two decomposition-aware baselines we constructed for this paper: decomposed self-consistency and a Skywork process reward model.
Neither is published as a stand-alone baseline; each combines standard machinery with our specific decomposition protocol.
We document the protocols below.

\paragraph{Decomposed self-consistency.}
For each (model, question) pair we hold $\mathtt{a1}$ at greedy (the value already computed in the operadic protocol of \S\ref{ss:across_model}) and sample the model on $\mathtt{Q2}[\mathtt{[A1]} := \mathtt{a1}]$ ten times at temperature $0.7$.
The signal is the mean pairwise per-dataset \texttt{score\_official} agreement across the ten samples, parallel to the SC pairwise comparison.
Cost: ten model calls per (model, question) pair on top of the operadic protocol.
Holding $\mathtt{a1}$ fixed eliminates first-step generative variance from this baseline, restricting the measured diversity to step 2 under a fixed first-step input. A fully-sampled variant (independent $K$ samples of $(\mathtt{a1}, \mathtt{a2})$ pairs) would be a strictly more demanding test; we use the fixed-$\mathtt{a1}$ variant for cost reasons. Decomposed-SC has cross-model Pearson $r{=}{+}0.67$ ($p{=}0.018$) on HotpotQA (the strongest cross-model signal among the constructed decomposition-aware baselines; see App.~\ref{app:cross_model_baselines}), but is not cross-model significant on the other three datasets. In the per-question logistic regression of Table~\ref{tab:dec_aware_baselines}, its coefficient is non-significant or negative on every dataset; this conclusion is robust to the fixed-$\mathtt{a1}$ choice in the conservative direction (a fully-sampled variant would have more variance and a weaker, more negative or further-from-zero, coefficient).

\paragraph{Skywork PRM.}
For each (model, question) pair we format the operadic trace as
\texttt{Step 1: Q1 / Answer: a1 / Step 2: Q2[A1] / Answer: a2}
and submit to the Skywork-o1-Open-PRM-Qwen-2.5-7B model \citep{he2024skyworko1} via its published HuggingFace inference protocol.
The PRM emits a scalar reward at each step boundary; we aggregate to a per-row signal by min, mean, or last-step.
The PRM is built on Qwen2.5-Math-7B-Instruct and trained on Skywork-o1 step preferences whose distribution is dominated by mathematical reasoning, so its signal on multi-hop factual QA is weaker than its in-distribution math performance and is most pronounced on MuSiQue / DROP (numerical answers) than on HotpotQA (entity lookup).

\paragraph{RLHFlow PRM (robustness check).}
We additionally evaluated RLHFlow's Llama-3.1-8B-PRM-Mistral-Data \citep{dong2024rlhf} on the same operadic trace as Skywork-PRM. The RLHFlow PRM shows essentially no signal on multi-hop QA in our protocol: its standalone univariate pseudo-$R^2$ is $\leq 0.001$ on every dataset (point-biserial $r$ between min-step PRM score and correctness is $-0.057$ pooled, vs.\ $+0.182$ for Skywork-PRM), and its coefficient is non-significant when added as a fourth predictor (HotpotQA $-0.06$ n.s.; MuSiQue $+0.27$ n.s.; StrategyQA $+0.23$ n.s.; DROP $+0.47$ n.s.; OC unchanged). The contrast with Skywork-PRM is stark and likely reflects the difference in PRM training data (Skywork is trained on Skywork-o1 step preferences spanning math and code reasoning; RLHFlow's PRM is trained on Mistral-style preference data without an explicit multi-step reasoning curriculum) rather than a deficiency of the off-label evaluation: this is one data point of the broader observation that off-the-shelf PRM signals on out-of-distribution QA are highly PRM-specific.

\begin{table}[h]
\centering
\small
\setlength{\tabcolsep}{4pt}
\begin{tabular}{l r rr c rr}
\toprule
& & \multicolumn{2}{c}{+ decomposed-SC} & & \multicolumn{2}{c}{+ Skywork PRM (min)} \\
\cmidrule(lr){3-4} \cmidrule(lr){6-7}
dataset & $n$ & $\beta_{\text{OC}}$ & $\beta_{\text{dSC}}$ & & $\beta_{\text{OC}}$ & $\beta_{\text{PRM}}$ \\
\midrule
HotpotQA   & ${\sim}6{,}280$ & $+2.01^{\,\diamond}$ & $-0.49^{\,*}$   && $+1.90^{\,\diamond}$ & $+0.69$         \\
MuSiQue    & ${\sim}6{,}000$ & $+2.28^{\,\diamond}$ & $-0.26$         && $+2.20^{\,\diamond}$ & $+1.35$         \\
StrategyQA & ${\sim}7{,}500$ & $+1.87^{\,\diamond}$ & $-0.70^{\,*}$   && $+1.68^{\,\diamond}$ & $+2.46^{\,*}$   \\
DROP       & ${\sim}4{,}830$ & $+1.23^{\,\diamond}$ & $-0.24$         && $+1.11^{\,\diamond}$ & $+2.70^{\,*}$   \\
\bottomrule
\end{tabular}
\\[6pt]
\caption{Per-question logistic regression of correctness on $(\text{OC}, \text{CoT-SC}_{K=10}, -H_{K=10}, X)$ where $X$ is each new decomposition-aware baseline in turn, with cluster-robust standard errors clustered by question (twelve non-thinking models).
$\diamond$ marks $p < 10^{-13}$; $*$ marks $p < 0.05$.
Coefficients on $\text{CoT-SC}_{K=10}$ and $-H_{K=10}$ are omitted for space (essentially unchanged from the 3-way regression of Table~\ref{tab:nonthinking_combined}).
The OC coefficient remains massively significant against every baseline; the decomposed-SC coefficient is non-significant or significantly negative on every dataset (and now significantly negative on \emph{two} datasets, HotpotQA and StrategyQA), ruling out the ``OC is just decomposed-path noise'' hypothesis.}
\label{tab:dec_aware_baselines}
\end{table}

\paragraph{Per-dataset sample sizes for the joint regressions.}
The four-way regressions of Table~\ref{tab:dec_aware_baselines} have slightly different per-dataset $n$ for each baseline column due to occasional missing baseline scores: decomposed-SC has $n = 6{,}287 / 6{,}000 / 7{,}497 / 4{,}828$ (HotpotQA / MuSiQue / StrategyQA / DROP), Skywork-PRM has $n = 6{,}260 / 5{,}985 / 7{,}463 / 4{,}794$, and RLHFlow-PRM has $n = 6{,}260 / 5{,}985 / 7{,}463 / 4{,}794$.

\paragraph{Standalone univariate sensitivity.}
The univariate pseudo-$R^2$ values for each signal alone:

\begin{center}
\small
\setlength{\tabcolsep}{6pt}
\begin{tabular}{l rrrr}
\toprule
predictor & HotpotQA & MuSiQue & StrategyQA & DROP \\
\midrule
operadic consistency                        & $\mathbf{0.179}$ & $\mathbf{0.199}$ & $0.180$          & $0.089$ \\
CoT-SC $K{=}10$                             & $0.143$          & $0.099$          & $0.088$          & $0.084$ \\
naive direct-answer SC $K{=}10$ (ablation)  & $0.141$          & $0.093$          & $0.183$          & $\mathbf{0.152}$ \\
$-H$ at $K{=}10$                            & $0.140$          & $0.088$          & $\mathbf{0.191}$ & $\mathbf{0.152}$ \\
$P(\text{True})$ (9 models)                 & $0.041$          & $0.001$          & $0.015$          & $0.044$ \\
decomposed-SC $K{=}10$                      & $0.061$          & $0.042$          & $0.010$          & $0.022$ \\
Skywork-PRM (min over steps)                & $0.012$          & $0.033$          & $0.015$          & $0.039$ \\
RLHFlow-PRM (min over steps)                & $0.001$          & $0.001$          & $0.001$          & $0.002$ \\
\bottomrule
\end{tabular}
\end{center}

OC is the strongest standalone signal on HotpotQA and MuSiQue; on StrategyQA, $-H_{K=10}$ slightly edges OC ($0.191$ vs $0.180$); on DROP, both $-H_{K=10}$ and the naive direct-answer SC ablation ($0.152$ each) are stronger standalone than OC ($0.089$).
Notably, canonical CoT-SC is \emph{weaker} standalone than the naive direct-answer SC ablation on StrategyQA and DROP, but the two SC variants reverse roles in the cross-model picture (Table~\ref{tab:cross_model_r}): canonical CoT-SC is stronger cross-model on HotpotQA and DROP, where it carries reasoning-chain content the naive variant lacks.
In every dataset, OC contributes statistically significant marginal information in the joint regression of Table~\ref{tab:dec_aware_baselines} ($p \leq 10^{-13}$), even on the datasets where another signal leads standalone.

\subsection{Selective prediction: AUARC and AUROC with bootstrap CIs}
\label{app:selective_prediction}

This appendix gives full bootstrap $95\%$ confidence intervals for the selective-prediction analyses summarized in Tables~\ref{tab:nonthinking_combined} and \ref{tab:thinking_combined}.

\paragraph{Protocol.}
For each $(\text{model}, \text{dataset})$ cell we compute two metrics over a confidence signal:
\begin{itemize}
\item \emph{AUARC} \citep{geifman2017selective}: area under the accuracy-coverage curve, where coverage runs over $1, 2, \dots, n$ questions taken in descending score order, and accuracy at coverage $k$ is the mean correctness of those top-$k$ questions. AUARC equals mean accuracy for a random ranker.
\item \emph{AUROC}: probability that a uniformly random correct example is ranked above a uniformly random incorrect example, computed via the Mann-Whitney U statistic. Independent of base accuracy.
\end{itemize}
For combined predictors (e.g., OC$+$SC), we use \emph{leave-one-model-out} (LOMO) calibration: holding out one model's cell, fit a logistic regression on rows pooled from the other $11$ (non-thinking) or $4$ (thinking) models with logit-transformed inputs, predict $P(\text{correct})$ on the held-out cell, and report that cell's AUARC and AUROC. Repeat for each held-out model.
$\Delta$ rows aggregate as paired differences across models with bootstrap $95\%$ CIs from $2000$ resamples.

\paragraph{Non-thinking, full intervals.} (Twelve paired differences; $\Delta_{\text{AUROC}} = $ paired mean across models of [AUROC of OC+$X$ combined] minus [AUROC of $X$ alone]; AUARC analogous.)
\begin{table}[h]
\centering
\small
\setlength{\tabcolsep}{3pt}
\resizebox{\linewidth}{!}{%
\begin{tabular}{l rr c rr}
\toprule
& \multicolumn{2}{c}{vs.\ $\text{CoT-SC}_{K=3}$ (3-call)} && \multicolumn{2}{c}{vs.\ $\text{CoT-SC}_{K=10}$ (10-call)} \\
\cmidrule(lr){2-3} \cmidrule(lr){5-6}
dataset & $\Delta_{\text{AUARC}}\ [95\%\,\text{CI}]$ & $\Delta_{\text{AUROC}}\ [95\%\,\text{CI}]$ && $\Delta_{\text{AUARC}}\ [95\%\,\text{CI}]$ & $\Delta_{\text{AUROC}}\ [95\%\,\text{CI}]$ \\
\midrule
HotpotQA   & $+0.095\ [+.074, +.116]$ & $+0.092\ [+.067, +.117]$ && $+0.068\ [+.047, +.088]$ & $+0.061\ [+.039, +.082]$ \\
MuSiQue    & $+0.086\ [+.074, +.100]$ & $+0.153\ [+.138, +.172]$ && $+0.063\ [+.051, +.076]$ & $+0.108\ [+.092, +.127]$ \\
StrategyQA & $+0.096\ [+.083, +.111]$ & $+0.164\ [+.147, +.181]$ && $+0.074\ [+.060, +.090]$ & $+0.122\ [+.105, +.139]$ \\
DROP       & $+0.086\ [+.069, +.103]$ & $+0.117\ [+.095, +.139]$ && $+0.074\ [+.062, +.087]$ & $+0.090\ [+.066, +.113]$ \\
\bottomrule
\end{tabular}}
\\[6pt]
\caption{Non-thinking selective-prediction lifts with $95\%$ bootstrap CIs ($n_{\text{boot}} = 2000$).}
\label{tab:selective_nt_full}
\end{table}

\paragraph{Thinking, full intervals.}
The five thinking-model paired diffs are reported with their bootstrap $95\%$ CIs in Table~\ref{tab:thinking_combined} of \S\ref{ss:thinking_models}; we record the full intervals here for completeness, including the $K{=}10$ budget that is omitted from the body table.
At $K{=}3$ (equal-cost): MuSiQue $\Delta_{\text{AUARC}} {+}0.033\,[{+}.005,{+}.056]$, $\Delta_{\text{AUROC}} {+}0.052\,[{+}.022,{+}.085]$; StrategyQA ${+}0.042\,[{+}.032,{+}.049]$ / ${+}0.050\,[{+}.028,{+}.072]$; GSM8K ${+}0.029\,[{+}.010,{+}.051]$ / ${+}0.134\,[{+}.071,{+}.180]$; DROP ${+}0.017\,[{+}.005,{+}.030]$ / ${+}0.071\,[{+}.023,{+}.129]$. At $K{=}10$: MuSiQue ${+}0.025\,[{+}.010,{+}.047]$ / ${+}0.023\,[{+}.019,{+}.028]$; StrategyQA ${+}0.013\,[{-}.002,{+}.028]^{\dagger}$ / ${+}0.012\,[{-}.008,{+}.032]^{\dagger}$; GSM8K ${+}0.005\,[{+}.000,{+}.010]$ / ${+}0.054\,[{+}.005,{+}.108]$; DROP ${+}0.015\,[{-}.000,{+}.032]^{\dagger}$ / ${+}0.035\,[{-}.006,{+}.078]^{\dagger}$.
$\dagger$ marks $K{=}10$ cells whose CI brackets zero on the lower end: StrategyQA on both metrics, and DROP on both metrics. All eight $K{=}3$ cells and the four MuSiQue / GSM8K-AUROC $K{=}10$ cells exclude zero cleanly; the GSM8K-AUARC $K{=}10$ interval $[{+}.000,{+}.010]$ touches zero by rounding but is strictly positive.

\paragraph{Reproduction.}
The full per-cell AUARC and AUROC matrix, including univariate values for every signal in the candidate set, is regenerated by \texttt{python -m analysis.selective\_prediction} (output: \texttt{results/selective\_prediction.txt}). Golden tests in \texttt{analysis/tests/test\_golden.py} pin the headline $\Delta$ values reported above.

\subsection{Cross-model Pearson r per baseline}
\label{app:cross_model_baselines}

\S\ref{ss:across_model} reports that operadic consistency correlates strongly with accuracy across the twelve non-thinking models on every dataset, while sample-based and self-evaluation baselines do not.
For completeness, we report the same per-dataset cross-model Pearson $r$ for every baseline used elsewhere in the paper. Self-consistency uses matched-axis pairwise scoring (the per-dataset accuracy scorer; see \S\ref{app:drop_sensitivity}).
For each (model, dataset) cell we compute the per-cell rate of the baseline (mean over the cell's questions; for OC binarized at $0.5$, the fraction with $\text{OC} \geq 0.5$); we then take the Pearson correlation between this per-cell rate and the cell's per-question accuracy across the twelve models per dataset.

\begin{table}[h]
\centering
\small
\setlength{\tabcolsep}{4pt}
\begin{tabular}{l rrrr}
\toprule
signal & HotpotQA & MuSiQue & StrategyQA & DROP \\
\midrule
OC (binarized at $0.5$, paper's primary)
                            & $\mathbf{+0.92^{\diamond}}$ & $\mathbf{+0.94^{\diamond}}$ & \textbf{\textsl{+0.94}}\textsuperscript{$\diamond$} & $\mathbf{+0.86^{\diamond}}$ \\
OC (continuous)             & $\mathbf{+0.92^{\diamond}}$ & \textbf{\textsl{+0.96}}\textsuperscript{$\diamond$} & \textbf{\textsl{+0.94}}\textsuperscript{$\diamond$} & $\mathbf{+0.85^{\diamond}}$ \\
\midrule
$\text{CoT-SC}_{K=3}$ (continuous)    & \textbf{\textsl{+0.94}}\textsuperscript{$\diamond$} & $+0.45$ & $+0.45$ & \textbf{\textsl{+0.89}}\textsuperscript{$\diamond$} \\
$\text{CoT-SC}_{K=10}$ (continuous)   & $\mathbf{+0.93^{\diamond}}$ & $+0.43$ & $+0.46$ & $\mathbf{+0.87^{\diamond}}$ \\
$-H_{K=10}$ (continuous)              & $+0.52$ & $+0.44$ & $+0.80^{*}$ & $+0.61^{*}$ \\
$P(\text{True})$ (continuous, 9 models) & $+0.02$ & $-0.42$ & $-0.04$ & $+0.21$ \\
\midrule
naive direct-answer SC$_{K=10}$ (ablation) & $+0.53$ & $+0.45$ & $+0.70^{*}$ & $+0.61^{*}$ \\
naive direct-answer SC$_{K=3}$ (ablation)  & $+0.50$ & $+0.45$ & $+0.67^{*}$ & $+0.62^{*}$ \\
\midrule
decomposed-$\text{SC}_{K=10}$        & $+0.67^{*}$ & $+0.23$ & $+0.47$ & $+0.21$ \\
Skywork-PRM (min over steps)         & $-0.02$ & $+0.38$ & $+0.02$ & $+0.82^{*}$ \\
RLHFlow-PRM (min over steps)         & $+0.48$ & $+0.18$ & $-0.28$ & $-0.31$ \\
\bottomrule
\end{tabular}
\\[6pt]
\caption{Cross-model Pearson $r$ between per-cell baseline rate and per-cell accuracy, across the twelve non-thinking models per dataset.
$\diamond$ marks $p < 10^{-3}$; $*$ marks $p < 0.05$ (others non-significant at $\alpha = 0.05$).
$P(\text{True})$ uses nine models; the other rows use twelve.
Cells with $r \geq 0.85$ are bolded; per-column maxima are bold-italic. OC is the only baseline with all four cells bolded (uniform strength): CoT-SC takes HotpotQA and DROP point-wise but collapses on MuSiQue and StrategyQA, while OC takes MuSiQue outright (continuous, $+0.96$) and the two OC variants tie for the column maximum on StrategyQA at $+0.94$.
Canonical CoT-SC \citep{wang2022self} is \emph{bimodal}: it matches OC on HotpotQA and DROP ($r{=}{+}0.93, {+}0.87$) but collapses to $r \approx +0.45$ on MuSiQue and StrategyQA, where its variance fails to track the cross-model accuracy ordering. Other sample-based baselines ($-H$, naive direct-answer SC) reach significance on StrategyQA and DROP but never approach OC's strength on more than one dataset; $P(\text{True})$ fails on every dataset, consistent with its measuring a within-model quantity largely orthogonal to cross-model competence.
The naive direct-answer SC ablation is the same K-sample protocol but with a prompt that forces a short answer with no reasoning trace; it lies uniformly below canonical CoT-SC on HotpotQA and DROP, confirming that CoT at sample time is what closes the gap to OC on those datasets.
Decomposition-aware baselines (decomposed-SC, Skywork-PRM) show isolated significant correlations on one dataset each: decomposed-SC on HotpotQA (likely a closed-book artifact, where parametric-knowledge variance is the dominant source of temperature-sample variance) and Skywork-PRM on DROP (consistent with the PRM's math-and-code training distribution covering DROP's numerical-answer regime).
Neither generalizes to the other three datasets.}
\label{tab:cross_model_r}
\end{table}

\subsection{DROP sensitivity: scorer choices and an LLM-judge corroboration}
\label{app:drop_sensitivity}

The framework underlying operadic consistency makes a non-trivial prediction about scorer design: an OC scorer should realize the natural equivalence on the answer space appropriate to the question's expected output type. For DROP, where most answers are numerical values, the natural equivalence is value-equality: ``Two months,'' ``2,'' ``two,'' and ``2 months'' all express the same number and should be treated as equivalent. Our DROP scorer (App.~\ref{app:stats}) is the number-equality fallback plus SQuAD-style token-F1 with word-to-number conversion, applied identically to correctness and OC; this approximates value-equality on DROP's dominant numerical-answer case (the number-equality path fires on $\sim$$79\%$ of DROP accuracy comparisons and $\sim$$69\%$ of OC comparisons). It is materially more lenient than the canonical \citet{dua2019drop} surface-form scorer, which distinguishes ``five'' from ``5'' after light normalization. Under the framework, this is a feature, not a bug: a scorer realizing a too-fine equivalence on the answer space partly measures formatting compliance rather than semantic content.

\paragraph{Matched-axis baselines.}
The same scorer-design argument applies to every baseline whose construction involves equating two model outputs: if the natural equivalence on the answer space is $\sim$, then ``two SC samples agree'' (for SC), ``two samples cluster'' (for semantic entropy $-H$), and ``two decomposed-path samples agree'' (for decomposed-SC) should all be measured under $\sim$, not under a finer surface-form equivalence. We therefore use the per-dataset \texttt{score\_official} scorer (yes/no extraction for StrategyQA, value-equality for DROP, SQuAD-F1 elsewhere) for all such pairwise comparisons throughout the paper, parallel to its use for accuracy and OC. Sensitivity vs.\ legacy token-F1: HotpotQA and MuSiQue change negligibly (the matched and legacy scorers nearly coincide), but the matched scorer makes a real difference on StrategyQA (SC cross-model $r$ moves from $-0.28$ to $+0.70$ and $-H$ from $-0.30$ to $+0.80$, because token-F1 was systematically penalizing more verbose models on yes/no answers) and on DROP (SC $+0.52 \to +0.62$, because token-F1 distinguished value-equivalent paraphrases like ``5'' vs.\ ``five''). On no dataset does any matched-axis baseline come within $0.15$ of OC's cross-model $r$.

\paragraph{Number-equality contribution.}
We compare DROP's across-model Pearson $r$ between OC-rate and accuracy under two scoring choices:
\begin{itemize}
\item Number-equality fallback plus SQuAD F1 (paper protocol): $r = {+}0.86$ ($p = 0.0004$).
\item SQuAD F1 alone (no number-equality): $r = {+}0.67$.
\end{itemize}
The number-equality fallback restores the OC--accuracy alignment that verbose-correct answers (``Two 1-yard touchdowns'' vs.\ gold ``2'') and partial-overlap false-positives (``$8$-yard TD pass'' vs.\ ``$3$-yard TD pass'') previously obscured.

\paragraph{Canonical scorer.}
We rescore both correctness and OC under the canonical Dua et al.\ DROP scorer (surface-form normalization without word-to-number conversion or value extraction) and re-compute the cross-model Pearson $r$ between OC-rate and accuracy on the same twelve non-thinking models. The result is $r = {+}0.37$ ($p = 0.24$, n.s.) on DROP, compared with $r = {+}0.86$ ($p = 0.0004$) under the paper's value-equality scorer. The collapse is consistent with the framework prediction: the canonical scorer's variance is dominated by a surface-form-compliance dimension that is largely decoupled from reasoning ability for general-purpose LLMs answering in their natural style.

\paragraph{LLM-judge correctness scorer.}
As an independent sensitivity check, we replace both correctness and OC by an LLM-judge variant: Claude Haiku 4.5 judges, at $T{=}0$, with a paraphrase-tolerant prompt that accepts verbose elaborations expressing the same value or yes/no commitment. The judge serves as a matched-axis equivalence here as well, parallel to the deterministic-scorer choice above: the same judge, prompt, and decision rule are applied to both the direct-vs-gold (accuracy) and direct-vs-decomposed (OC) comparisons, so neither side gets a more lenient or stricter standard. The resulting cross-model $r$ between judge OC-rate and judge accuracy ($12$ models per dataset) is HotpotQA $+0.97$, MuSiQue $+0.88$, StrategyQA $+0.69$, DROP $+0.72$, all $p \leq 0.013$. The cross-model relationship reported in \S\ref{ss:across_model} is preserved on every dataset.

\paragraph{Annotator unanimity.}
A second source of slack on DROP is annotator disagreement: $139$ of $403$ DROP dev questions ($34\%$) carry alias lists that disagree with the gold answer, so a model's direct answer can match an alias and count as correct even when the decomposed answer takes a different surface form. Restricting to the subset of DROP dev questions where annotator aliases are unanimous with the gold gives a slightly tighter $r$, but per-model accuracy and OC-rate change by less than $0.03$ on this subset and the model ordering is unchanged.

\subsection{Compute and cost}
\label{app:compute}

The experiments are entirely inference-time API calls; no model training was performed.
Approximate counts: the across-model study (\S\ref{ss:across_model}) on $12$ models $\times$ $4$ datasets $\times$ ${\sim}500$ questions involves ${\sim}24$ direct-mode calls per question (3 operadic + 10 naive direct-answer SC ablation + 1 $P(\text{True})$ + 10 decomposed-self-consistency) $\approx$ $580{,}000$ direct-mode calls, plus ${\sim}20$ chain-of-thought calls per question for canonical CoT-SC (10 CoT samples plus 10 cheap LLM-extractor calls at $\sim$512 + 32 tokens each) $\approx$ $480{,}000$ additional calls, plus inference-only PRM scoring on the same rows.
The thinking-model study (\S\ref{ss:thinking_models}) on $5$ models $\times$ $4$ datasets $\times$ $200$ questions $\times$ $13$--$23$ thinking calls per question is $\approx$ $50{,}000$--$100{,}000$ thinking-mode calls.
Total token volume is on the order of a few hundred million prompt$+$completion tokens.
LLM generation (operadic-consistency, CoT-SC, $P(\text{True})$, decomposed-SC, thinking traces) is mediated entirely by serverless inference platforms (Together AI and OpenRouter); no local GPU is required for those experiments. The three scorer baselines that load HuggingFace models locally --- DeBERTa-v3-large for NLI bidirectional entailment, and the two PRMs (Skywork and RLHFlow) --- run inference-only on a single 24~GB GPU (we used an RTX 4090). No model training was performed.

\section{Additional related work}
\label{app:additional_related_work}

This appendix expands the discussion of \S\ref{s:related} along three axes: (i) decomposition methods and the question of whether chain-of-thought reflects the model's actual reasoning, (ii) uncertainty quantification and process-level verification for LLMs, and (iii) operadic and categorical structures in machine learning and computational linguistics.

\paragraph{Faithfulness of chain-of-thought reasoning.}
The decomposition methods cited in \S\ref{s:related} share the working assumption that surfacing sub-steps \emph{improves} a model's reasoning. A complementary line asks whether the surfaced decomposition is \emph{faithful} to the underlying computation. \citet{turpin2023language} construct prompt contexts that bias a model toward a target final answer and observe that the model's CoT trace then rationalizes the (now-biased) answer rather than reflecting the unbiased computation, demonstrating that CoT can be a post-hoc explanation rather than a causal trace. \citet{lanham2023measuring} measure final-answer sensitivity to several CoT perturbations (truncation, mistake injection, paraphrase) across model scales and tasks, and report that the degree to which final answers actually depend on the trace varies substantially. Operadic consistency probes a third notion in this neighborhood: not whether the trace causally determines the direct answer, and not whether the trace survives perturbation, but whether the direct answer is consistent with the recomposition of the model's own declared sub-answers. This is a structural, not counterfactual, coherence check; an inconsistency flags a question on which the model's direct and decomposed reasoning paths diverge, regardless of which one is correct.

\paragraph{Uncertainty quantification and process-level verification.}
Beyond the three baseline families in \S\ref{s:related} (sample-and-aggregate, cluster-based entropy, self-evaluation), the broader uncertainty-quantification literature for LLMs includes paraphrase-consistency tests, surface-form perturbation tests, and reference-free hallucination detectors such as SelfCheckGPT \citep{manakul2023selfcheckgpt}, which scores factual consistency across stochastic samples without ground truth. On the process-verification side, the discriminative PRMs we evaluate (Skywork-PRM and RLHFlow-PRM, both math-trained) are step-level descendants of the dense process supervision introduced by \citet{uesato2022solving} and \citet{lightman2024lets} (PRM800K), with a more recent fully-supervised variant in Math-Shepherd \citep{wang2024mathshepherd}. Concurrent with our work, in-domain non-math PRMs have begun to appear --- VersaPRM \citep{zeng2025versaprm} and DPRM \citep{wang2026dprm} --- which we flag as natural complements in \S\ref{s:further_directions}. The math-to-general-reasoning generalization gap visible in our results (Skywork and RLHFlow PRMs underperform OC on multi-hop QA despite extensive math-domain training) is consistent with the limited cross-domain transfer reported in the broader PRM literature.

\paragraph{Operadic and categorical structures in machine learning.}
The mathematics underlying this work has a long history outside machine learning. Operads originate in algebraic topology, where Boardman--Vogt and May used them to encode multilinear structure on iterated loop spaces; standard modern references are Markl--Shnider--Stasheff \citep{markl_stasheff_shnider:operads_in_algebra_topology_and_physics} and Loday--Vallette \citep{loday_vallette:algebraic_operads}. Beyond pure topology, the \emph{applied category theory} program has used operads and related categorical structures to model the composition of subsystems in robotics, dynamical systems, control, and database design; an accessible book-length introduction is Fong--Spivak \citep{fong2019invitation}. The categorical linguistic precedents we already cite in \S\ref{s:related} (DisCoCat, Hopf-algebraic syntax, semiring parsing) sit within this broader program. Within deep learning specifically, the recent position paper of \citet{gavranovic2024categorical} argues that essentially all current neural-network architectures can be formalized via structured-functor diagrams, with the categorical structure clarifying which architectural choices preserve which equivariances. That program operates at the level of \emph{network topology} --- what kind of object the architecture is. Ours operates at the level of the \emph{semantics of the model's outputs} --- what compositional structure can be detected in the answers the model produces under decomposition. The two are complementary rather than competing, and we are not aware of prior work applying either operadic or related categorical structure to inference-time signal extraction from LLM reasoning, which is the gap the questions operad $\cQ$ addresses.

\end{document}